\documentclass[letterpaper]{article} % DO NOT CHANGE THIS
\usepackage{aaai25}  % DO NOT CHANGE THIS
\usepackage{times}  % DO NOT CHANGE THIS
\usepackage{helvet}  % DO NOT CHANGE THIS
\usepackage{courier}  % DO NOT CHANGE THIS
\usepackage[hyphens]{url}  % DO NOT CHANGE THIS
\usepackage{graphicx} % DO NOT CHANGE THIS
\usepackage{natbib}  % DO NOT CHANGE THIS AND DO NOT ADD ANY OPTIONS TO IT
\usepackage{caption} % DO NOT CHANGE THIS AND DO NOT ADD ANY OPTIONS TO IT
\usepackage{algorithm}
\usepackage{algorithmic}
\usepackage{multirow}
\usepackage{amsmath}
\usepackage{amssymb}
\usepackage{amsthm}
\usepackage{booktabs}
\usepackage{tabularx}
\usepackage{appendix}
\usepackage{enumitem}

\usepackage{newfloat}
\usepackage{listings}
\DeclareCaptionStyle{ruled}{labelfont=normalfont,labelsep=colon,strut=off} % DO NOT CHANGE THIS
\floatstyle{ruled}
\newfloat{listing}{tb}{lst}{}
\floatname{listing}{Listing}
\title{Federated Graph Condensation with Information Bottleneck Principles}
\author{
    Bo Yan\textsuperscript{\rm 1,2},
    Sihao He\textsuperscript{\rm 1},
    Cheng Yang\textsuperscript{\rm 1},
    Shang Liu\textsuperscript{\rm 3},
    Yang Cao\textsuperscript{\rm 2},
    Chuan Shi\textsuperscript{\rm 1}\thanks{Corresponding author.}
}

\affiliations{
    \textsuperscript{\rm 1}Beijing University of Posts and Telecommunications\\
    
    \textsuperscript{\rm 2}Institute of Science Tokyo\\
    
    \textsuperscript{\rm 3}China University of Mining and Technology\\
      \{boyan, sihaohe, yangcheng, shichuan\}@bupt.edu.cn,
    shang@cumt.edu.cn,
    cao@c.titech.ac.jp\\

}

\usepackage{bibentry}
\begin{document}

\maketitle
% \vspace*{-15mm}
\begin{abstract}
Graph condensation (GC), which reduces the size of a large-scale graph by synthesizing a small-scale condensed graph as its substitution, has benefited various graph learning tasks. However, existing GC methods rely on centralized data storage, which is unfeasible for real-world decentralized data distribution, and overlook data holders' privacy-preserving requirements. To bridge this gap, we propose and study the novel problem of federated graph condensation (FGC) for graph neural networks (GNNs). Specifically, we first propose a general framework for FGC, where we decouple the typical gradient matching process for GC into client-side gradient calculation and server-side gradient matching, integrating knowledge from multiple clients' subgraphs into one smaller condensed graph. Nevertheless, our empirical studies show that under the federated setting, the condensed graph will consistently leak data membership privacy, i.e., the condensed graph during federated training can be utilized to steal training data under the membership inference attack (MIA). To tackle this issue, we innovatively incorporate information bottleneck principles into the FGC, which only needs to extract partial node features in one local pre-training step and utilize the features during federated training. Theoretical and experimental analyses demonstrate that our framework consistently protects membership privacy during training. Meanwhile, it can achieve comparable and even superior performance against existing centralized GC and federated graph learning (FGL) methods.
\end{abstract}


\section{Introduction}

\begin{figure}[t]
    % \vspace*{-13mm}
    \centering
    \includegraphics[scale=.55]{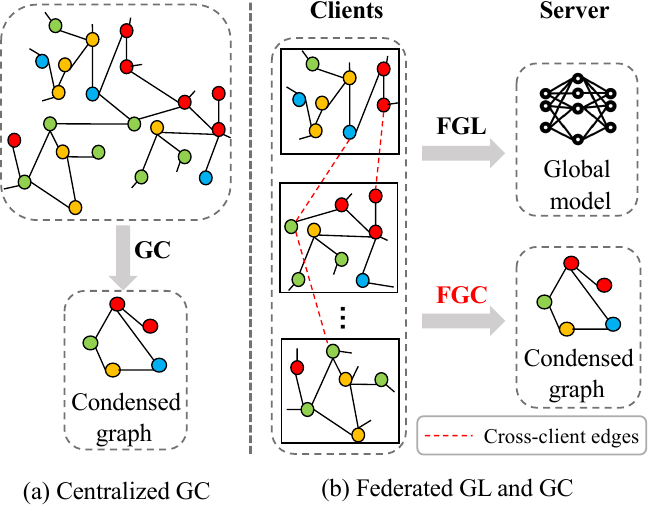}
    \vspace{-2mm}
    \caption{The comparison between (a) centralized graph condensation(GC), (b) federated graph learning (FGL) and federated graph condensation (FGC).}
    \label{fig:gc}
\vspace{-5mm}
\end{figure}

Graph data, such as social networks, transportation networks, and molecular interactions, is ubiquitous in the real world. Graph neural networks (GNNs) have been prevalent for modeling these graphs and achieved remarkable success in recent years \cite{DBLP:conf/nips/HamiltonYL17, DBLP:conf/iclr/KipfW17,DBLP:conf/iclr/VelickovicCCRLB18}. However, GNNs have struggled with real-world large-scale graphs due to memory and computational overheads, which are further exacerbated when the models need to be trained multiple times, e.g., neural architecture search (NAS) \cite{DBLP:conf/cvpr/ZhangPCSHH022} and continual learning \cite{DBLP:conf/eccv/LiH16}. A natural solution from the data-centric view is graph condensation (GC) \cite{DBLP:conf/iclr/JinZZLTS22, DBLP:conf/kdd/JinTJLZTY22, DBLP:journals/corr/abs-2206-13697, DBLP:conf/nips/ZhengZCN0P23}, as shown in Figure \ref{fig:gc}(a), which reduces large-scale graphs into small ones but keeps the utility of original graphs, facilitating the storage of graph data and computations of downstream tasks. As a typical technique of GC, gradient matching \cite{DBLP:conf/iclr/JinZZLTS22, DBLP:conf/kdd/JinTJLZTY22} forces the gradients of GNN models trained on the original graph and condensed graph to be the same, maintaining a comparable model performance, but largely reducing the graph size.

Despite significant progress in GC, existing works all hold a basic assumption that the graph data is centrally stored. In reality, the whole graph is divided into multiple subgraphs held by different parties. For example, in healthcare systems, a hospital only possesses a subset of patients (nodes), their information (attributes), and interactions (links). Due to privacy concerns, data holders are unwilling to share their data and many strict regulations such as GDPR \cite{voigt2017eu} also forbid collecting data arbitrarily, which makes centralized GC unfeasible. Federated graph learning (FGL) \cite{DBLP:conf/nips/YaoJRJ23, DBLP:conf/mlsys/WanLLKL22, DBLP:conf/nips/ZhangYLSY21} has emerged in recent years to endow collaboratively training a global model without exposing local data, as is shown in Figure \ref{fig:gc}(b). Typically, they are dedicated to training a global model by exchanging encrypted information for recovering missing structures (e.g., cross-client edges and neighbors). However, they still suffer from burdensome computations of traditional GNNs for large-scale graph training, especially in the cross-silo scenario \cite{DBLP:journals/corr/abs-2210-04505}. For example, if the NAS for FGL is conducted, it inevitably performs federated training multiple times, which exacerbates both computations and communications. 
Therefore, an important yet unexplored area lies in considering GC in the federated setting.

To bridge this gap, we investigate the novel problem of federated graph condensations (FGC), as shown in Figure \ref{fig:gc} (b),  which aims to collaboratively learn a small graph that contains knowledge from different data sources. Subsequently, the condensed graph can be accessed by all data holders for downstream applications, such as node classifications. This task is non-trivial due to two significant challenges. \textbf{\textit{(1) How to preserve the utility of the condensed graph that is learned from distributed subgraphs?}} Unlike centralized GC the graph data is integral, the entire graph is distributed as subgraphs across data holders in FGC. As a result, some critical cross-client information is missing and these subgraphs also suffer from severe heterogeneity, which inevitably degrades the utility of the condensed graph. In this regard, a unified framework for utility-preserving FGC is desired. \textbf{\textit{(2) How to preserve the local graph's membership privacy that may be leaked by the condensed graph?}} Membership privacy is widely used to measure information leakage about training data \cite{DBLP:conf/sp/NasrSH19}. Unlike local storage of graph data in typical FGL, a condensed graph is released in FGC, which may be utilized to infer the membership privacy of local data using membership inference attack (MIA). Furthermore, the accessibility of the condensed graph enables attackers to train arbitrary models, as shown in Figure \ref{fig:mia}, rendering traditional model-based defenses against MIA infeasible. Thus, new methods are needed to defend against MIA in FGC.

To tackle these challenges, we endeavor to condense graphs in the federated setting and propose a \underline{Fed}erated  \underline{G}raph \underline{C}ondensation framework (FedGC). (1) To preserve data utility, we propose a general framework for FGC, the core of which is that the condensed graph is learned by matching the aggregated gradients from clients. Particularly, the gradients are aggregated in a class-aware weighted manner to tackle data heterogeneity. Furthermore, an encrypted one-step communication between the server and clients is adopted to recover the cross-client neighbors. During the evaluation phase, the model trained on the condensed graph will be fine-tuned by local data for better personalization. (2) To preserve membership privacy, we further propose a novel local graph transformation module with information bottleneck principles \cite{DBLP:conf/iclr/AlemiFD017}. We first empirically reveal that the condensed graph will consistently leak membership privacy. To address this, each subgraph will be transformed locally before condensation by fixing the graph structure unchanged and extracting sufficient but minimal node features. Then the transformed graph will act as a proxy of the original graph to conduct FGC, thereby protecting the local membership privacy. Besides, to alleviate the performance decay in graph transformation caused by scarce labeled nodes and heterogeneous feature space, we propose a one-time self-training strategy to label nodes and utilize a fixed pre-trained shared model to align the feature space. Finally, we give theoretical and empirical analysis to prove the utility- and privacy-preserving properties of FedGC. The major contributions of this paper are summarized as follows:

\begin{itemize}
  \item 
  
  To the best of our knowledge, this is the first work to study federated graph condensation (FGC), which is an important and practical task in real-world scenarios. 
  
  \item We design a FedGC framework for FGC. It learns the condensed graph by matching the weighted gradient aggregation from clients. We reveal that the condensed graph will consistently leak membership privacy and further propose a novel local graph transformation module with information bottleneck principles to protect original graphs. Theoretical analysis proves that FedGC can simultaneously maintain the utility of the original graph and preserve membership privacy.
  
  % It leverages rich semantic information of HIN by performing meta-path-guided neighbor aggregations. 
  \item We conduct extensive experiments on five real-world datasets and show that FedGC outperforms centralized GC and FGL methods, especially in large-scale datasets. Meanwhile, FedGC can consistently protect membership privacy during the whole federated training process.
\end{itemize}

\section{Preliminary}
\label{preliminary}

\noindent \textbf{Notations}. Let $G=(A,X,Y)$ denote the original graph with adjacency matrix $A \in \mathbb{R}^{N \times N}$, node feature matrix $X \in \mathbb{R}^{N \times d}$ and node label set $Y \in \{1,\ldots,C\}$ over $C$ classes. Similarly, let $G^{\prime}=(A^{\prime},X^{\prime},Y^{\prime})$ denote the condensed graph. In the federated setting, the whole graph $G$ is divided into multiple subgraphs $\{G_{i}\}_{i=1}^{M}$ where $G_{i} = (A_{i},X_{i},Y_{i})$ stored in the $i$-th client. Let $V$ denote the whole node set and $V_i$ denote the node set of the $i$-th client, following \cite{DBLP:conf/nips/YaoJRJ23}, we assume $V=V_1 \cup \ldots  \cup V_M$ and $V_i \cap V_j=\emptyset$, i.e., there are no overlapping nodes across clients. Also, we use $E$ to denote the whole edge set and $E_i$ to denote the edge set of $i$-th client. For an edge $e_{v,u} \in E$, where $v \in V_i$ and $u \in V_j$, we assume $e_{v,u} \in E_i \cup E_j$, i.e., the cross-client edge $e_{v,u}$ is known to the client $i$ and $j$.

\begin{figure}[t]
    \centering
    \includegraphics[scale=.45]{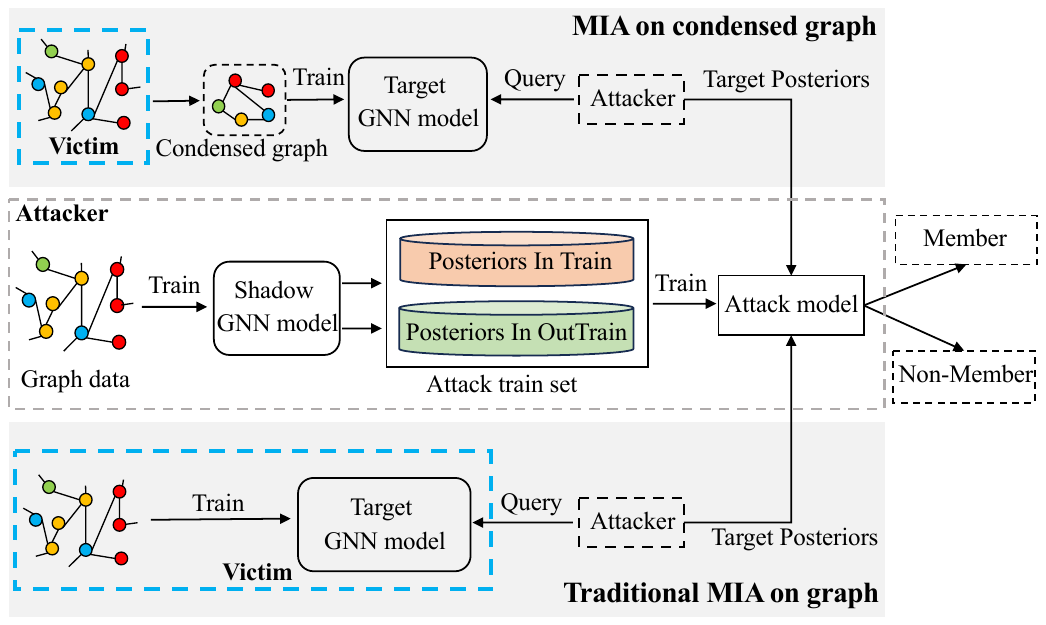}
    \caption{The comparison between MIA on condensed graphs and traditional MIA on graphs. The components circled by blue dashed lines are invisible to attackers.}
    \label{fig:mia}
\vspace{-3mm}
\end{figure}

\noindent \textbf{Information bottleneck} \cite{DBLP:conf/iclr/AlemiFD017}. Given original data $\mathcal{D}$, information bottleneck (IB) is to optimize $Z$ to capture the minimal sufficient information within  $\mathcal{D}$ to predict the label $Y$. The objective of IB can be written as:

\vspace{-3mm}
\begin{equation} 
\small
\min -\text{I}(Y; Z) + \gamma \text{I}(\mathcal{D}; Z)
\label{eq:pre_ib},
\vspace{-1mm}
\end{equation}
where $\text{I}(;)$ means mutual information (MI).
Maximizing the first term aims to preserve the utility of $\mathcal{D}$ and minimizing the second term helps filter out the redundant information of $\mathcal{D}$. $\gamma$ is the parameter that balances two terms.

\noindent \textbf{GC with gradient matching}. 
Given a graph $G$, GC aims to condense $G$ into a smaller, synthetic graph $G^{\prime}=(A^{\prime},X^{\prime},Y^{\prime})$ with $A^{\prime} \in \mathbb{R}^{N^{\prime} \times N^{\prime}}$, node feature matrix $X^{\prime} \in \mathbb{R}^{N^{\prime} \times d}$, node label set $Y^{\prime} \in \{1,\ldots,C\}^{N^{\prime}}$ and $N^{\prime} \ll N$, meanwhile maintaining the utility of $G$ (i.e., achieving comparable performance with the model trained on $G$). To achieve this, typical solutions update $G^{\prime}$ by matching the gradients $\nabla \mathcal{L}$ of models trained on $G$ and $G^{\prime}$ \cite{DBLP:conf/iclr/ZhaoMB21}, which can be formalized as:

\vspace{-2mm}
\begin{equation} 
\small
    \min_{G^{\prime}} E_{\theta_0 \sim P_{\theta}} [D(\nabla_{\theta}\mathcal{L}^{G}_{\theta_0}, \nabla_{\theta}\mathcal{L}^{G^{\prime}}_{\theta_0})],
\label{eq:doscond}
\vspace{-1mm}
\end{equation}
where $D(\cdot,\cdot)$ is the distance function (e.g., Euclidean distance) and $\theta_0$ is the initialized model parameter.

\noindent \textbf{MIA on condensed graphs}. 
Following \cite{DBLP:conf/icml/DongZL22}, we assume an honest-but-curious server is the strong attacker. Since the condensed graph $G^{\prime}$ is publicly accessed, the target models can be arbitrarily selected and trained by attacks, which is different from traditional MIA on graphs (the attacker can only query target models \cite{DBLP:conf/tpsisa/OlatunjiNK21}), as shown in Figure \ref{fig:mia}. Therefore, traditional model-based defenses against MIA are infeasible. Following \cite{DBLP:conf/tpsisa/OlatunjiNK21}, we also assume the attacker knows the target data distribution and can access a part of in-distribution data $G_s$ (shadow data). To conduct MIA in FGC scenarios, the attacker first utilizes $G_s$ to train a shadow GNN $\mathcal{S}$. Then, the posterior probabilities of nodes from $\mathcal{S}$ are used to train an attack model $\mathcal{A}$ to distinguish between non-member and member nodes. Meanwhile, the attacker utilizes $G^{\prime}$ to train a target GNN $\mathcal{T}$. During the attack phase, given a target node $v \in V$ and its $L$-hop neighbors, the posterior probability of $v$ will be obtained by feeding $v$ into $\mathcal{T}$. Then, $\mathcal{A}$ can infer the membership of $v$ based on its posterior probability. Note that the $L$-hop neighbors of $v$ known to the attacker may not be the exact neighbors, i.e., the attacker would not know the exact local structure, which is also a realistic setting.

\noindent \textbf{Task formulation}. Given multiple subgraphs $\{G_{i}=(A_{i},X_{i},Y_{i})\}_{i=1}^{M}$, each of which stores in the $i$-th client, FGC aims to condense $\{G_{i}\}_{i=1}^{M}$ into a smaller graph $G^{\prime}$ stored in the server. $G^{\prime}$ should maintain the utilities of original graphs $\{G_{i}\}_{i=1}^{M}$ (i.e., the model trained on $G^{\prime}$ should obtain the comparable performance to that of a model trained on the original graphs), meanwhile containing less membership privacy of original graphs.

% \begin{figure*}[t]
%     \centering
%     \includegraphics[scale=.9]{figures/framework.pdf}
%     \caption{The general framework of FedGC.}
%     \label{fig:framework}
% \end{figure*}

\section{Methodology}
In this section, we give a detailed introduction to the proposed model FedGC. We first present a general framework for FGC. Then, we empirically reveal that the condensed graph by the framework is vulnerable to MIA. Thus, a local graph transformation with information bottleneck principles is introduced to defend MIA. Finally, we theoretically prove the utility- and privacy-preserving abilities of FedGC. The algorithm of FedGC can be found in the Appendix.

\subsection{General Framework of FGC}
\label{sec:gf_fgc}
We first review the typical centralized graph condensation in Eq.(\ref{eq:doscond}). It can not be directly applied to the federated setting since the integral graph $G=(A,X,Y)$ is divided into multiple subgraphs $\{G_j=(A_j, X_j, Y_j)\}_{j=1}^{M}$. A natural solution is that each client $j$ trains a GNN model locally based on $G_j$ and uploads model gradients to the server for aggregation. To mitigate the impact of node label heterogeneity among different clients and facilitate optimization, for each class $c$, we first sample a subgraph $G_{j,c}=(A_{j,c}, X_{j,c}, Y_{j,c})$ consisting of nodes with label $c$ and their neighbors, then calculate the model gradients on $G_{j,c}$. The server performs weighted gradient aggregation based on the number of nodes $n_j(c)$. Therefore, we can modify the matching loss in Eq.(\ref{eq:doscond}) as:

\vspace{-4mm}
\begin{equation} 
    \small
    \begin{split}
    \min_{G^{\prime}} \frac{1}{t}\sum^{t=T}_{t=0} \sum_{c=0}^{C}[D(\nabla_{\theta}\mathcal{L}^{G}_{c,\theta_t}, \nabla_{\theta}\mathcal{L}^{G^{\prime}}_{c,\theta_t})],
    \theta_t \sim P_{\theta},\\
    \nabla_{\theta}\mathcal{L}^{G}_{c,\theta_t}=\sum_{j=1}^{j=M}\frac{n_j(c)}{n(c)} \nabla_{\theta}\mathcal{L}(GNN_{\theta_t}(A_{j,c},X_{j,c}),Y_{j,c}), \\
    \nabla_{\theta}\mathcal{L}^{G^{\prime}}_{\theta_t,c} = \nabla_{\theta}\mathcal{L}(GNN_{\theta_t}(A^{\prime}_c,X^{\prime}_c),Y^{\prime}_c).    
    \end{split}
    \label{eq:fedgc1}
\vspace{-4mm}
\end{equation}
That is, at each communication round $t$, the condensed graph can be updated by matching the gradient $\nabla_{\theta}\mathcal{L}^{G}_{c, \theta_t}$ and $\nabla_{\theta}\mathcal{L}^{G^{\prime}}_{c,\theta_t}$ w.r.t. GNN on current condensed graph $G^{\prime}$. Note that we leave the updating of the condensed graph on the server side and the client only needs to upload the gradient of the GNN model, which reduces the communication cost and computation overhead of clients.

As illustrated in \cite{DBLP:conf/iclr/JinZZLTS22}, it's hard to optimize $A^{\prime}$, $X^{\prime}$ and $Y^{\prime}$ jointly. Therefore, we fix the $Y^{\prime}$ as the same distribution as the original $Y$ during federated training. To obtain the distribution $P(Y)$, the client $j$ will upload the number of each class $n_j(c)$ in its own $Y_j$ before federated training. We also assume that $A^{\prime}$ is induced from $X^{\prime}$: $A_{i,j}^{\prime}=\sigma(\text{MLP}_{\Phi}(x_i^{\prime}, x_j^{\prime}))$, which largely reduces computation cost.

% For updating $A^{\prime}$ and $X^{\prime}$, we model their correlations during optimizing by assuming the graph structure $A^{\prime}$ is induced from node features $X^{\prime}$: $A_{i,j}^{\prime}=\sigma(\text{MLP}_{\Phi}(x_i^{\prime}, x_j^{\prime}))$, which largely reduces computation cost. 

\noindent \textbf{One-step communication}.
The challenges in FGL lie in broken local structures and the distributional heterogeneity of subgraphs \cite{DBLP:conf/nips/YaoJRJ23, DBLP:conf/icml/BaekJJYH23}, which are further compounded by the utility-preserving property in FGC. Therefore, inspired by \cite{DBLP:conf/nips/YaoJRJ23}, we leverage homomorphic encryption \cite{DBLP:journals/toct/BrakerskiGV14} to securely transfer missing neighbor aggregations among clients. This step is performed only once in advance, and the final aggregated node embeddings are used for FGC. Notably, as at most 2-hop neighbors are sufficient in cross-silo FL \cite{DBLP:conf/nips/YaoJRJ23}, this step introduces little communication cost while improving performance. 

\noindent \textbf{Fine-tuning with local graphs}.
After the condensed graph $G^{\prime}$ is well trained, it can be distributed to clients for testing. A typical testing flow in traditional GC is training a target model $\mathcal{T}$ with $G^{\prime}$, then utilizing the model $\mathcal{T}$ to test local data. However, in FGC, the heterogeneity issue across clients hinders the performance of this one-test-all mode since the condensed graph only captures the general knowledge of multiple subgraphs. Therefore, we propose to use local data to fine-tune $\mathcal{T}$ so that it can adapt well to different data distributions. 

\begin{figure}[t]
    \centering
    \includegraphics[scale=.095]{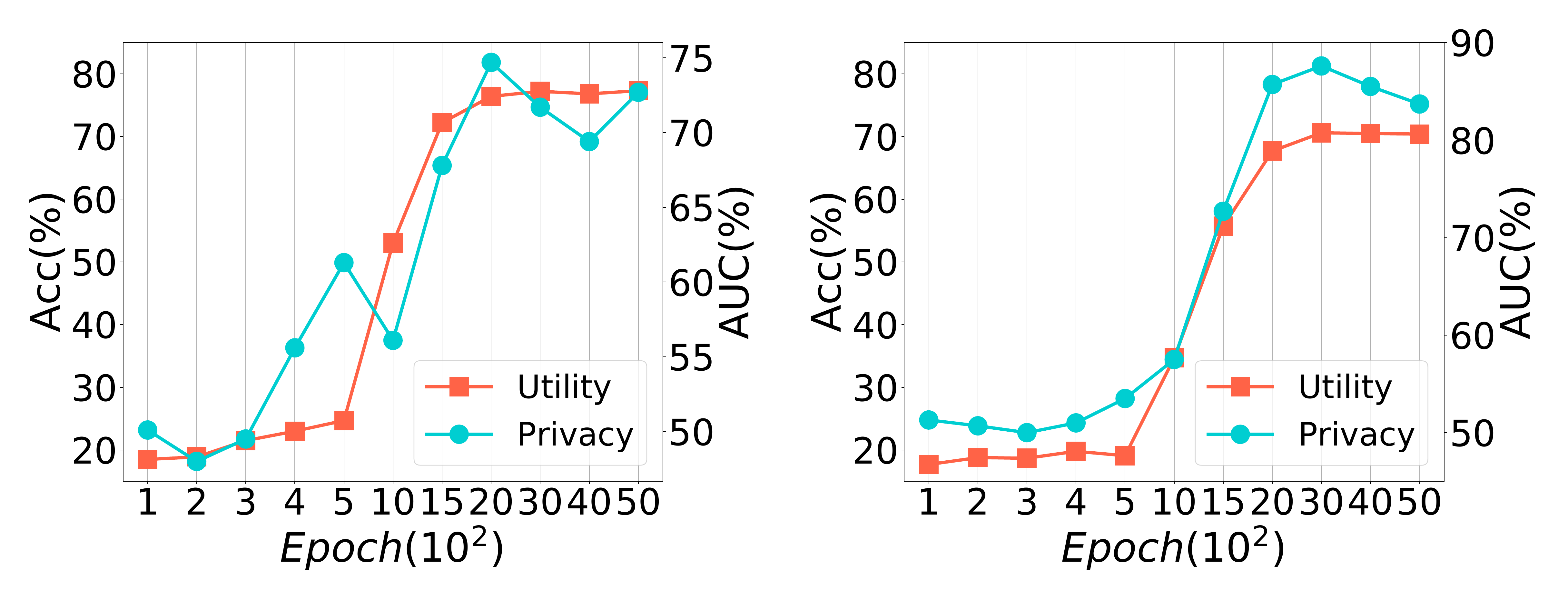}
    \vspace{-3mm}
    \caption{The comparison between the utility of condensed graph (accuracy of node classification $\uparrow$) and privacy attack performance (AUC of MIA $\downarrow$) during federated training (left: Cora, right: Citeseer).}
    \label{fig:motivation}
\vspace{-4mm}
\end{figure}

\subsection{Local Graph Transformation for Preserving Membership Privacy}
\label{sec:ib}
% \subsection{Empirical Study for Membership Privacy in Federated Graph Condensation}
% \label{sec:motivation}
Based on the above framework, we can condense multiple large subgraphs into a small graph without exposing local data. However, our empirical study shows that the condensed graph still exposes membership privacy. We perform FGC under the MIA introduced in the preliminary to record the utility and attack performance of the condensed graph. From Fig. \ref{fig:motivation}, we can see that the performance of MIA increases along with the utility, i.e., the condensed graph will gradually leak the membership privacy of original graphs, which intuitively makes sense since it's trained to match the local graphs. Hereby, new mechanisms are desired to condense graphs containing less membership privacy. 

\begin{figure}[t]
    \centering
    \includegraphics[scale=0.85]{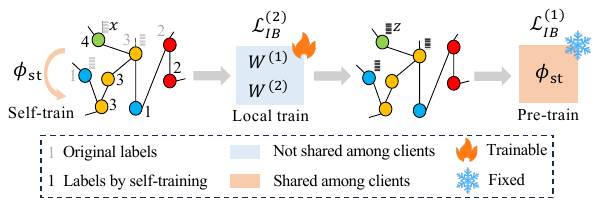}
    \caption{The overall workflow of local graph transformation with information bottleneck principles.}
    \label{fig:gib}
\vspace{-5mm}
\end{figure}

To achieve this, we propose to transform the local subgraph into another one, and then perform FGC based on these transformed graphs. The new graph should be equipped with two properties: (1) maintaining the utility of the original graph and (2) containing less membership privacy of the original graph. Inspired by recent work \cite{DBLP:conf/nips/YangZZ0PL023, DBLP:conf/kdd/DaiCWTWCYW23} that utilizes information bottleneck (IB) principles to extract minimal sufficient features from origin data, we propose to transform original local subgraphs with IB principles. The overall workflow is depicted in Fig. \ref{fig:gib}. Concretely, for a node in one client, we use $x \in \mathbb{R}^{d \times 1}$ to denote its original features. Then we aim to learn an IB encoder $f_x$ to extract minimal sufficient features $z$ of $x$:

\vspace{-3.5mm}
\begin{equation} 
    z = f_x(x, \mathcal{N}),
    \label{eq:ibe}
\end{equation}
where $\mathcal{N}$ is the neighborhood of the node. According to the IB principles \cite{DBLP:conf/iclr/AlemiFD017,DBLP:conf/nips/WuRLL20}, our objective function is:

\vspace{-3mm}
\begin{equation} 
\small
    \min -\text{I}(z,\mathcal{N};y)+\gamma \text{I}(z;x,\mathcal{N}).
    \label{eq:ib}
\end{equation}
By optimizing Eq.(\ref{eq:ib}), the first term $\text{I}(z,\mathcal{N};y)$ is maximized, which makes $z$ contain enough information to predict the label $y$. Meanwhile, the second term $\text{I}(z;x,\mathcal{N})$ is minimized, forcing $z$ to learn less information from $x$ and their neighbors $\mathcal{N}$. As a result, the new feature $z$ can preserve the utility and contains less privacy information of original features $x$.

\noindent \textbf{Upper bound of the objective function}.  It's hard to directly optimize Eq.(\ref{eq:ib}) due to the two MI terms. Following the IB principles \cite{DBLP:conf/iclr/AlemiFD017}, we derive the upper bound of the two MI terms. First, we introduce a variational distribution $q(y|z,\mathcal{N})$ to approximate $p(y|z,\mathcal{N})$, and we derive the upper bound of $-\text{I}(z,\mathcal{N};y)$:

\vspace{-5mm}
\begin{equation}
\small
\begin{aligned}
-\text{I}(z,\mathcal{N};y) 
&\leq -\mathbb{E}_{p(z,\mathcal{N},y)}\log \frac{q(y|z,\mathcal{N})}{p(y)}\\
% &=-\mathbb{E}_{p(z,\mathcal{N},y)} \log q(y|z,\mathcal{N})-H(y)\\
&\leq -\mathbb{E}_{p(z,\mathcal{N},y)} \log q(y|z,\mathcal{N})=\mathcal{L}_{IB}^{(1)}.
\end{aligned}
\label{eq:izy2}
\end{equation}
Similarly, a variational distribution $q(z)$ is introduced to approximate $p(z)$, then the upper bound of $\text{I}(z;x,\mathcal{N})$ can be calculated by: 

\vspace{-4mm}
\begin{equation}
\small
\begin{aligned}
\text{I}(z;x,\mathcal{N})
&\leq \mathbb{E}_{p(z,x,\mathcal{N})} \log \frac{p(z|x,\mathcal{N})}{q(z)}\\
&=\mathbb{E}_{p(x,\mathcal{N})}\text{KL}(p(z|x,\mathcal{N}) || q(z))=\mathcal{L}_{IB}^{(2)}.
\end{aligned}
\label{eq:izx2}
\end{equation}
After obtaining the above upper bounds, we get the optimization objective of our local graph transformation: 

\vspace{-1mm}
\begin{equation}
\small
\min \mathcal{L}_{IB} = \mathcal{L}_{IB}^{(1)} + \gamma \mathcal{L}_{IB}^{(2)}.
\label{eq:gt}
\end{equation}

It should be noted that the typical graph information bottleneck \cite{DBLP:conf/nips/WuRLL20} also performs graph structure transformation. Considering the joint optimization of features and structures brings burdensome computation to the clients, we only perform node feature transformation and keep the original structure unchanged, which can be further theoretically proven to defend against MIA.

\noindent \textbf{Instantiating the IB Principle}. After giving the above IB principles, we need to instantiate them with different neural network architectures. For $p(z|x,\mathcal{N})$, we assume it follows Gaussian distribution, which is parameterized by MLPs followed by neighbor aggregations. Specifically, we first feed $x$ into a two-layer MLP: $h = W^{(2)}\sigma(W^{(1)}x)$, where $W^{(1)} \in \mathbb{R}^{d \times d}$ and $W^{(2)} \in \mathbb{R}^{2d \times d}$. Then, we perform local neighbor aggregation to obtain hidden representation $\hat{z} = \text{AGGREGATE}(h, \mathcal{N})$. $z$ is sampled from the Gaussian distribution with mean and variance from $\hat{z}$:

\vspace{-4mm}
\begin{equation}
\small
z \sim N(\mu;\sigma), \;\; \mu = \hat{z}[0:d],\;\; \sigma=\text{softplus}(\hat{z}[d:2d]).
\label{eq:pzxn}
\vspace{-1mm}
\end{equation}
In this process, the reparameterization trick \cite{DBLP:journals/corr/KingmaW13} is applied. We assume $q(z)$ follows the standard normal distribution $\mathcal{N}(0,1)$, then the $\mathcal{L}_{IB}^{(2)}$ can be optimized by minimizing the distance (e.g., KL divergence) between $q(z)$ and $p(z|x,\mathcal{N})$. After obtaining $z$, we utilize vanilla GCN \cite{DBLP:conf/iclr/KipfW17} with $\phi$ to parameterize $q(y|z,\mathcal{N})$. Thus, the $\mathcal{L}_{IB}^{(1)}$ can be instantiated as follows for node classification task:

\vspace{-3mm}
\begin{equation}
\small
\mathcal{L}_{IB}^{(1)} = \mathbb{E}_{p(z,\mathcal{N},y)} \text{Cross-Entropy}(\text{GCN}_{\phi}(z,\mathcal{N}), y).
\label{eq:ib1}
\end{equation}

\noindent \textbf{Federated training with transformed graphs}. For each client $j$, we can obtain a new subgraph $G^t_j=(A_j, Z_j, Y_j)$ after local graph transformation. The $Z_j$ with the best validation performance is stored locally. Then, the client utilizes $G^t_j$ to perform FGC using the general framework in Section \ref{sec:gf_fgc}. However, there exist two major flaws: (1) lack of enough labeled samples to obtain high-quality $Z_j$. Since each client only owns a subgraph of the entire graph, the labeled nodes in each client are scarce, making it hard to apply IB principles \cite{DBLP:conf/kdd/DaiCWTWCYW23}. (2) the obtained $Z$ of different clients are not in the same feature space. To better protect privacy, the graph transformation is conducted locally, i.e., there is no parameter sharing across clients, leading to a heterogeneous feature space and undermining the subsequent FGC. To address the first flaw, we propose a federated self-train strategy to expand labeled node sets. Before local graph transformation, all clients first collaboratively train a GNN model $\phi_{st}$ based on Fedavg \cite{DBLP:conf/aistats/McMahanMRHA17}. Then, each client utilizes $\phi_{st}$ to label local unlabeled nodes. To address the second flaw, we propose to initialize the same $\phi$ among clients with pre-trained $\phi_{st}$ and fix it during federated training. Each client only needs to locally update $W^{(1)}$ and $W^{(2)}$. In this way, $Z$ in different clients are forced to be unified in the same space.

\subsection{Privacy and Utility Analysis}
We theoretically analyze the privacy- and utility-preserving properties of FedGC. Let $\mathcal{G}^{\prime}=(A^{\prime},X^{\prime})$, $\mathcal{G}_i=(A_i,X_i)$, $\mathcal{G}_i^t=(A_i,Z_i)$ denote condensed graph, client $i$'s original graph and transformed graph respectively. Assuming $Y_i \rightarrow   \mathcal{G}_i \rightarrow \mathcal{G}_i^t \rightarrow \mathcal{G}^{\prime}$ form a Markov chain, the attack aims to find a graph $g$ to match $\mathcal{G}_i$ with an inference cost $C(\mathcal{G}_i, g)$.

\noindent \textbf{Definition 1.} \textit{Inference cost gain} \cite{DBLP:conf/allerton/CalmonF12}. Inference cost gain measures the quantitative improvement of inferring private data $S$ after observing $R$, denoted as $\Delta C = c^*_S - \mathbb{E}_{P_R}\left[c^*_R\right]$, where $c^*_S$ and $c^*_R$ is the minimum average cost of infer $S$ w/o and w/ $R$.  

In our case, the attack aims to infer $\mathcal{G}_i$ after observing $\mathcal{G}^{\prime}$. Therefore, we should minimize $\Delta C = c^*_{\mathcal{G}_i} - \mathbb{E}_{P_{\mathcal{G}^{\prime}}}\left[c^*_{\mathcal{G}^{\prime}}\right]$. Inspired by \cite{DBLP:conf/itw/MakhdoumiSFM14}, we have:

\noindent \textbf{Lemma 1.} If $L = \sup|C(\mathcal{G}_i, g)| < \infty$ is satisfied, then $\Delta C = c_{\mathcal{G}_i}^* - \mathbb{E}_{P_{\mathcal{G}^{\prime}}} \left[ c_{\mathcal{G}^{\prime}}^* \right] \leq 2\sqrt{2}L \sqrt{\text{I}(\mathcal{G}_i; \mathcal{G}^{\prime} )}$.

Lemma 1 demonstrates that $\Delta C$ is upperbounded by a function of $\text{I}(\mathcal{G}_i; \mathcal{G}^{\prime})$ under any bounded $C$.

\noindent \textbf{Lemma 2.} Minimizing $\text{I}(\mathcal{G}_i; \mathcal{G}_i^t)$ is equivalent to minimizing $\text{I}(A_i, X_i; Z_i)$.

Thus, given that $\text{I}(\mathcal{G}_i; \mathcal{G}^{\prime}) \leq \text{I}(\mathcal{G}_i; \mathcal{G}_i^t)$ deduced from the Markov chain, we have the following proposition.

\noindent \textbf{Proposition 1.} By minimizing $\text{I}(\mathcal{G}_i; \mathcal{G}_i^t)$, the \textit{inference cost gain} is also minimized.

Therefore, minimizing $\text{I}(A_i, X_i; Z_i)$ could defend the MIA theoretically based on the definition of \textit{inference cost gain}. As for utility, since $\mathcal{G}^{\prime}$ stems from $\mathcal{G}_i^t$, we can directly maximum the utility of $\mathcal{G}_i^t$, i.e., maximize $\text{I}(A_i, Z_i; Y_i)$. Noting that this maximization will increase the lower-bound of $\text{I}(A_i, X_i; Z_i)$ in light of $\text{I}(\mathcal{G}_i^t; Y_i) \leq \text{I}(\mathcal{G}_i^t; \mathcal{G}_i)$, thus the objective in Eq.(\ref{eq:ib}) can be seen as a trade-off between privacy- and utility-preserving of the condensed graph, which is controlled by $\gamma$. We will present this effect in the experiments. The proofs of these analyses can be found in the Appendix.

% It can be proved that under the privacy metric introduced in \cite{DBLP:conf/itw/MakhdoumiSFM14}, the objective in Eq.(\ref{eq:ib}) can be seen as a trade-off between privacy- and utility-preserving of the condensed graph. 
% the privacy leakage can be measured by the mutual information $I(z;x,\mathcal{N})$ and the utility of condensed graph can be measured by $I(z, \mathcal{N};y)$. 

% \noindent \textbf{Privacy}. In MIA, the attacker aims to perform an inference attack on the private graph data $X$, which can be formalized as finding a distribution $q \in \mathcal{P}_X$ to minimize the distribution cost function $C(X, q)$, then the inference cost is defined as:

% \vspace{-2mm}
% \begin{equation}
% c_{0}^{*}=\underset{q \in \mathcal{P}_{X}}{\min } \mathbb{E}_{P_{X}}[C(X, q)].
% \label{eq:q0}
% \end{equation}
% After observing the condensed graph $G$, the cost is modified as:

% \vspace{-3mm}
% \begin{equation}
% c_{g}^{*}=\underset{q \in \mathcal{P}_{X}}{\min } \mathbb{E}_{P_{X \mid G}}[C(X, q) \mid G=g].
% \label{eq:qy}
% \end{equation}
% Then the privacy leakage can be denoted as 

\section{Experiments}
\subsection{Experimental Setup}

\begin{table*}[t]
\footnotesize
\setlength\tabcolsep{4pt}
\center
\begin{tabular}{@{}c c ccc ccc c c@{}}
\toprule
\multirow{3}{*}{Dataset}  & \multirow{3}{*}{Ratio ($r$)}  & \multicolumn{3}{c}{Graph Condensation (GC)} & \multicolumn{3}{c}{Federated Graph Learning (FGL)} & \multirow{3}{*}{\begin{tabular}[c]{@{}c@{}}FedGC\end{tabular}} & \multirow{3}{*}{\begin{tabular}[c]{@{}c@{}}GCN\end{tabular}} \\
\cmidrule(l{2pt}r{2pt}){3-5} \cmidrule(l{2pt}r{2pt}){6-8}
        &  & GCOND & DOSCOND & SFGC & FedGCN & FedSage+ & BDS-GCN & &  \\ \midrule
\multirow{3}{*}{Cora}   
& 1.3\% & $79.8 {\pm}1.3$ & $80.0{\pm}1.4$ &$80.1{\pm}0.4$&  &  &  & \textbf{81.0${\pm}$1.1} & \\
& 2.6\% & $80.1{\pm}0.6$ & $79.2{\pm}0.1$ & \textbf{81.7${\pm}$0.5}&$80.6{\pm}0.4$  & $80.3{\pm}0.5$ & $76.0{\pm}1.4$  & $80.8{\pm}1.8$  & 81.2${\pm}$0.2 \\
& 5.2\% & $79.3{\pm}0.3$ & $79.3{\pm}1.3$& $81.6{\pm}0.8$& & & & \textbf{82.0${\pm}$0.7}  & \\ \midrule
\multirow{3}{*}{Citeseer}
& 0.9\% & $70.5{\pm}1.2$ & $71.4{\pm}0.2$ &$71.4{\pm}0.5$&  &  &  & \textbf{71.6${\pm}$0.1} & \\
& 1.8\% & $70.6{\pm}0.9$ & $70.5{\pm}0.5$ & \textbf{72.4${\pm}$0.4} &$69.3{\pm}0.7$  & $69.8{\pm}1.0$ & $67.1{\pm}1.8$  & ${70.5}{\pm}1.9$  & 71.7${\pm}$0.1 \\
& 3.6\% & $69.8{\pm}1.4$ & $70.5{\pm}1.1$& $70.6{\pm}0.7$& & & &\textbf{70.7${\pm}$1.0} & \\ \midrule
\multirow{3}{*}{Ogbn-arxiv}
& 0.05\% & $59.2{\pm}1.1$ & $59.1{\pm}0.8$ &$65.5{\pm}0.7$&  &  &  & \textbf{75.3${\pm}$0.8} & \\
&0.25\% & $63.2{\pm}0.3$ & $60.4{\pm}0.9$ & $66.1{\pm}0.4$ &$70.8{\pm}1.7$  & $70.5{\pm}0.5$ & $68.5{\pm}0.7$  & \textbf{75.8${\pm}$0.2} & 71.4${\pm}$0.1\\
&0.5\% & $64.0{\pm}0.4$ & $59.8{\pm}0.7$& $66.8{\pm}0.4$& & & & \textbf{75.3${\pm}$0.8}& \\ \midrule
\multirow{3}{*}{Flickr}
&0.1\% & $49.5{\pm}0.1$ & $48.0{\pm}0.1$ &$50.1{\pm}0.2$&  &  &  & \textbf{56.2${\pm}$2.4} & \\
&0.5\% & $49.7{\pm}0.1$ & $49.1{\pm}0.2$ & $49.5{\pm}0.2$ &$51.4{\pm}0.2$  & $51.2{\pm}0.1$ & $49.9{\pm}0.5$  & \textbf{55.1${\pm}$1.2} & 54.3${\pm}$0.1 \\
&1\% & $49.2{\pm}0.1$ & $49.1{\pm}0.1$& $50.5{\pm}0.1$& & & & \textbf{57.1${\pm}$1.1}& \\ \midrule
\multirow{3}{*}{Reddit}
& 0.05\% & $89.8{\pm}0.1$ & $89.4{\pm}0.6$ &$90.3{\pm}0.3$&  &  &  & \textbf{94.2${\pm}$0.2} & \\
&0.1\% & $89.8{\pm}0.2$ & $90.0{\pm}0.4$ & $91.2{\pm}0.4$ &$91.6{\pm}0.4$  & $90.7{\pm}0.4$ & $91.4{\pm}0.2$  & \textbf{94.3${\pm}$0.0}  & 94.5${\pm}$0.0\\
&0.2\% & $92.2{\pm}0.1$ & $91.6{\pm}0.5$& $92.1{\pm}0.1$& & & &\textbf{94.2${\pm}$0.2} & \\ \bottomrule 
\end{tabular}
\caption{Overall performance (\%) of condensed graphs by FedGC under different condensation rations compared with baselines.}
\label{tab:main}
\vspace{-2mm}
\end{table*}

\begin{table}[t]
\centering
\setlength\tabcolsep{1.5mm}
\footnotesize
\begin{tabular}
{@{}ccccccc@{}}
\toprule
\multicolumn{1}{l}{}                                  & Metrics & \textbf{--}    & PL  & Reg  & LDP    &  FedGC   \\ \midrule
Cora         & ACC $\uparrow$    & $79.4$      & $81.2$ & $79.1$ & $78.6$ & $80.8$ \\ 
$2.6\%$& AUC $\downarrow$     & $76.4$      & $63.3$ & $72.3$ & $71.4$ & $53.2$  \\
\midrule
Citeseer         & ACC $\uparrow$    & $71.3$      & $71.7$ & $71.2$ & $71.9$ & $70.5$ \\
$1.8\%$& AUC $\downarrow$   & $87.2$      & $68.4$ & $83.4$ & $85.1$ & $55.0$ \\
\midrule
Ogbn-arxiv & ACC $\uparrow$    & $73.2$      & $73.8$ & $72.5$ & $73.8$ & $75.8$ \\
$0.25\%$& AUC $\downarrow$    & $64.0$      & $60.2$ & $62.1$ & $63.2$ & $51.6$ \\
\midrule
Flickr       & ACC $\uparrow$    & $54.0$      & $55.5$ & $52.8$ & $55.2$ & $55.1$ \\
$0.5\%$& AUC $\downarrow$    & $62.1$      & $59.8$ & $60.7$ & $62.0$ & $54.8$  \\
\midrule
Reddit      & ACC $\uparrow$     & $93.1$       & $93.4$ & $92.0$ & $93.7$ & $94.3$ \\
$0.1\%$& AUC $\downarrow$     & $61.1$       & $57.3$ & $60.9$ & $61.2$ & $52.3$\\

\bottomrule

\end{tabular}
\caption{Overall performance (\%) of defending MIA by FedGC compared with baselines. \textbf{--} means removing IB.}
\label{tab:MIA}
\vspace{-5mm}
\end{table}

\textbf{Datasets}. Follow \cite{DBLP:conf/iclr/JinZZLTS22, DBLP:conf/nips/ZhengZCN0P23}, we evaluate FedGC on five graph datasets on node classification task, including Cora, Citeseer, Ogbn-arxiv, Flickr, and Reddit. We adopt the public splits provided in \cite{DBLP:conf/iclr/JinZZLTS22}. 

% Among them, Cora, Citeseer, and Ogbn-arxiv are citation networks, Flickr is a graph used for image categorization and Reddit is a large-scale graph of online discussion forum. 

\noindent \textbf{Baselines}. We choose three kinds of baselines to evaluate the effectiveness of condensed graphs by FedGC, including (1) \textit{centralized GCN} \cite{DBLP:journals/corr/KipfW16}, (2) \textit{centralized GC methods}: GCOND \cite{DBLP:conf/iclr/JinZZLTS22}, DosCond \cite{DBLP:conf/kdd/JinTJLZTY22} and SFGC \cite{DBLP:conf/nips/ZhengZCN0P23}, and (3) \textit{FGL methods}: BDS-GCN \cite{DBLP:conf/mlsys/WanLLKL22}, FedSage+ \cite{DBLP:conf/icml/BaekJJYH23} and FedGCN \cite{DBLP:conf/nips/YaoJRJ23}. Following \cite{DBLP:conf/kdd/DaiCWTWCYW23}, we replace IB principles with three defense methods to evaluate the effectiveness of FedGC in defending MIA, including PL \cite{lee2013pseudo}, Reg \cite{DBLP:conf/ccs/NasrSH18} and LDP \cite{DBLP:conf/isca/ChoiTVHK18}.

\noindent \textbf{Implementation}. Following \cite{DBLP:conf/iclr/JinZZLTS22}, we report results under different condensation ratio $r$.  Following \cite{DBLP:conf/nips/YaoJRJ23}, we test all FGL methods under the non-i.i.d. depicted by Dirichlet distribution ($\beta$=1). we set the client number $n$=10 for small datasets Cora and Citesser, and $n$=5 for large-scale datasets Ogbn-arxiv, Flickr, and Reddit. We run 5 times and report the average and variance of results. We utilize accuracy (Acc) to evaluate the condensation performance and AUC score to measure MIA performance. 

More details of setups can be found in the Appendix.

\begin{table}[t]
\centering
\setlength\tabcolsep{1mm}
\footnotesize
\begin{tabular}
{@{}cccccccccc@{}}
\toprule
\multicolumn{1}{l}{}                                                                 & Methods & MLP     & \multicolumn{1}{m{0.57cm}<{\centering}}{AP.}  & Cheby  & GCN    & SAGE   & SGC    & Avg.   \\ \midrule
\multirow{3}{*}{\begin{tabular}[c]{@{}c@{}}Cora\\       $2.6\%$\end{tabular}}    & GCOND    & $73.1$      & $78.5$ & $76.0$ & $80.1$ & $78.2$ & $79.3$ & $78.4$ \\ 
& SFGC     & $81.1$      & $78.8$ & $79.0$ & $81.1$ & $81.9$ & $79.1$ & $80.3$ \\
& \text{FedGC}     & $77.6$ & $81.2$ & $79.2$ & $80.8$ & $81.9$ & $81.6$ & $81.0$ \\\midrule
\multirow{3}{*}{\begin{tabular}[c]{@{}c@{}}Cite-\\seer\\      $1.8\%$\end{tabular}}         & GCOND    & $63.9$      & $69.6$ & $68.3$ & $70.5$ & $66.2$ & $70.3$ & $69.0$ \\
& SFGC    & $71.3$      & $70.5$ & $71.8$ & $71.6$ & $71.7$ & $71.8$ & $71.5$ \\
& \text{FedGC}  & $66.1$ & $71.2$ & $70.5$ & $70.5$ & $71.1$ & $70.2$ & $70.7$ \\\midrule
\multirow{3}{*}{\begin{tabular}[c]{@{}c@{}}Ogbn-\\arxiv\\       $0.25\%$\end{tabular}} & GCOND    & $62.2$      & $63.4$ & $54.9$ & $63.2$ & $62.6$ & $63.7$ & $61.6$ \\
& SFGC    & $65.1$      & $63.9$ & $60.7$ & $65.1$ & $64.8$ & $64.8$ & $64.3$ \\
& \text{FedGC} & $71.0$ & $75.9$ & $75.7$ & $75.8$ & $76.0$ & $75.0$ & $75.7$ \\
\bottomrule

\end{tabular}
\caption{Generalizability (\%) of condensed graphs by FedGC. \textbf{AP.}:APPNP. \textbf{Avg.}: the average test accuracy across different architectures (excluding MLP).}
\label{tab:generalization}
\vspace{-5mm}
\end{table}

\subsection{Overall Performance}
\noindent \textbf{Performance on condensed graph}. Table \ref{tab:main} shows the overall performance of the condensed graph compared with FGL, centralized GC, and GCN methods. We have the following observations: (1) FedGC achieves superior performance among all the baselines, demonstrating the effectiveness of our framework, which also sheds light on new ways to perform GC and FGL tasks. (2) Surprisingly, FedGC performs even better than FGL methods in the non-i.i.d. setting, given heterogeneity (non-i.i.d.) among subgraphs is a key challenge in FGL \cite{DBLP:conf/icml/BaekJJYH23}. We attribute this to the condensed graphs's ability to capture common knowledge across different subgraphs, thereby alleviating heterogeneity. Thus the trained model can be fine-tuned by local data to achieve better personalization. (3) FedGC achieves comparable results with centralized GCN and GC methods, especially in large-scale datasets. Compared to centralized GCN, the gains may lie in that the condensed graph captures general graph patterns that would not be affected by local noisy structures. As for GC methods, this superiority is mainly attributed to our elaborate designs, such as self-training and fine-tuning, which will be explicated in the ablation study.

\noindent \textbf{Performance on MIA}. We replace IB with other defense baselines to evaluate the effects of defending MIA. From Table \ref{tab:MIA}, we observe although the baselines can achieve comparable results with FedGC on the utility of condensed graphs, they suffer a dramatic performance decline in defending MIA. By contrast, FedGC can simultaneously achieve high utility and preserve membership privacy. Since the condensed graph is public, model-based defenses (Reg and LDP) can only affect the model gradients matching stage, which has little intervention on the properties entailed from original graphs, leading to a high success of MIA. In terms of this, PL directly modifies the original data (adding pseudo labels), thus obtaining better defense. However, the pseudo labels inherit the knowledge of training labels, making MIA still effective. 
Leveraging IB principles, FedGC transforms original node features into a new space, preventing strong attackers from concluding memberships.

%\end{enumerate}

\subsection{Generalizability of Condensed Graphs}
 Table \ref{tab:generalization} presents the generalizability of condensed graphs, i.e., we use the condensed graph to train different GNN models for testing local data.  Following \cite{DBLP:conf/iclr/JinZZLTS22}, we choose APPNP \cite{DBLP:conf/iclr/KlicperaBG19}, Cheby \cite{DBLP:conf/nips/DefferrardBV16}, GCN, GraphSAGE \cite{DBLP:conf/nips/HamiltonYL17} and SGC \cite{DBLP:conf/icml/WuSZFYW19} to report the average performance. We also include MLP for comparison. We can find that FedGC shows good generalizability across different GNN models. It outperforms GCOND in almost all datasets and achieves comparable and even superior performance than the SOTA method SFGC. The root cause is the similar filtering behaviors of these GNN models, which have been illustrated in studies \cite{DBLP:conf/iclr/JinZZLTS22, DBLP:conf/www/ZhuWSJ021}. The generalizability of condensed graphs enables clients to train personalized models based on self-conditions, which provides a flexible way to conduct downstream tasks. 

% \noindent \textbf{Different architectures}

% \noindent \textbf{Versatility of FedGC}

\begin{figure}[t]
    \centering
    \includegraphics[scale=0.23]{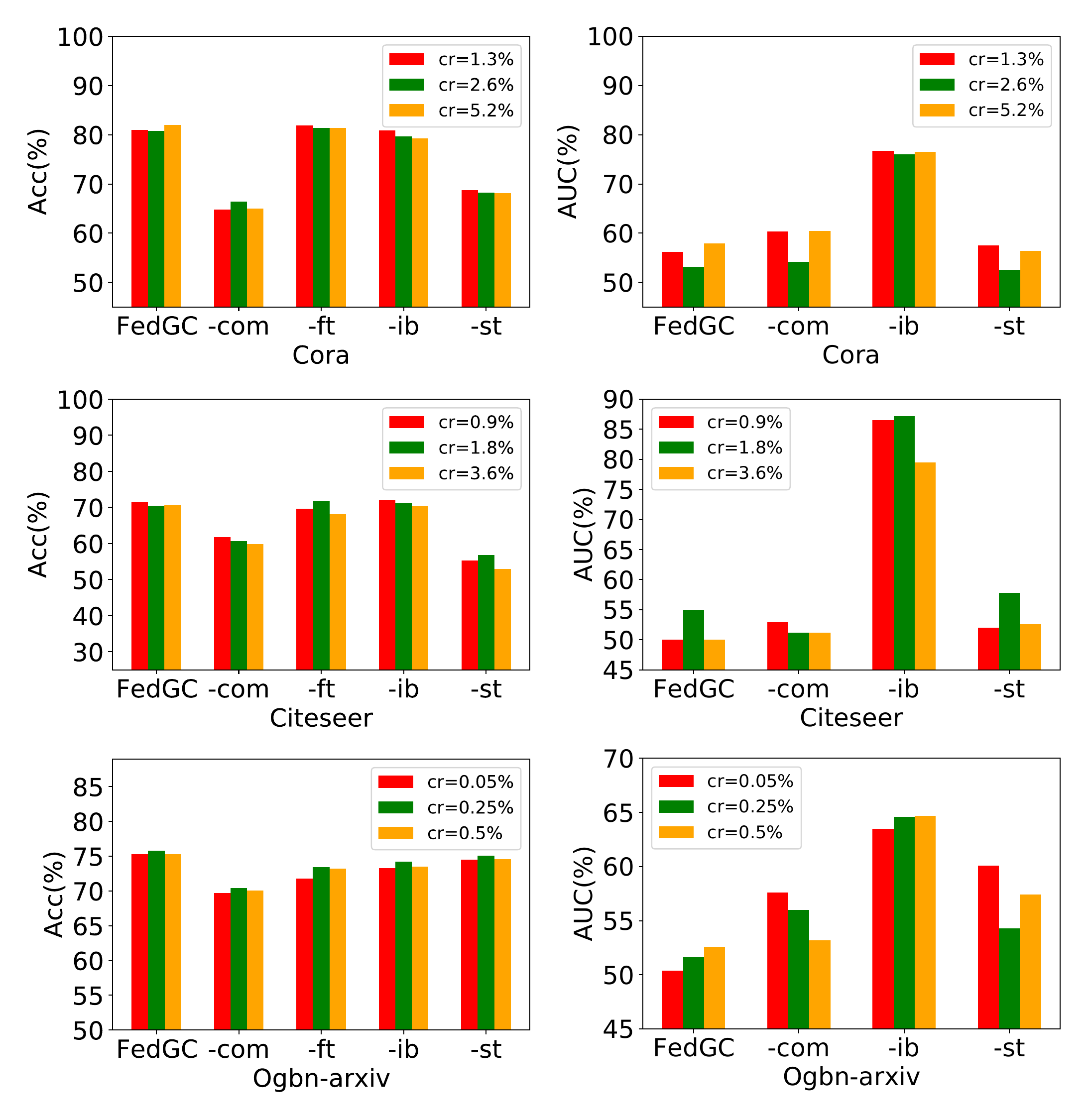}
    \vspace{-7mm}
    \caption{Ablation study for the performance of condensed graph (left $\uparrow$) and MIA (right $\downarrow$).}
    \label{fig:ab_performance}
\vspace{-4mm}
\end{figure}

%  \begin{figure*}[t]
%     \centering
%     \includegraphics[scale=0.245]{figures/ab_mia_performance.pdf}
%     \caption{Ablation study for the performance of MIA ($\downarrow$). }
%     \label{fig:ab_mia}
% \end{figure*}

 \begin{figure}[t]
    \centering
    \includegraphics[scale=0.095]{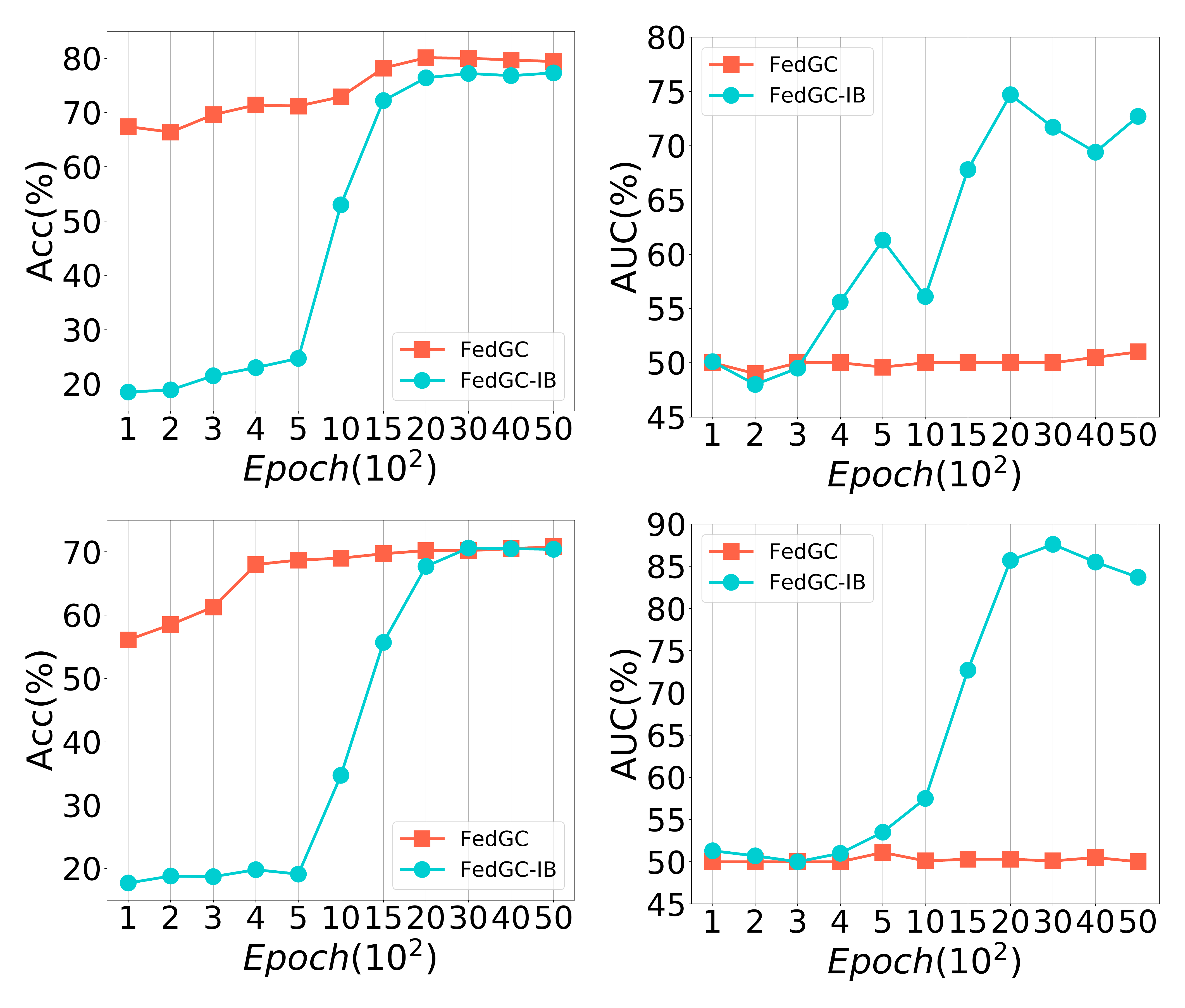}
    \vspace{-4mm}
    \caption{Training curve of condensed graphs (left $\uparrow$) and MIA (right $\downarrow$) w/o IB. (up: Cora, down: Citeseer).}
    \label{fig:ab_ib_performance}
\vspace{-3mm}
\end{figure}

%  \begin{figure}[t]
%     \centering
%     \includegraphics[scale=0.12]{figures/ab_train_mia_curve.pdf}
%     \caption{Training curve of MIA performance ($\downarrow$) w/o IB (left: Cora, right: Citeseer).}
%     \label{fig:ab_ib_mia}
% \end{figure}

 \begin{figure}[t]
    \centering
    \includegraphics[scale=0.095]{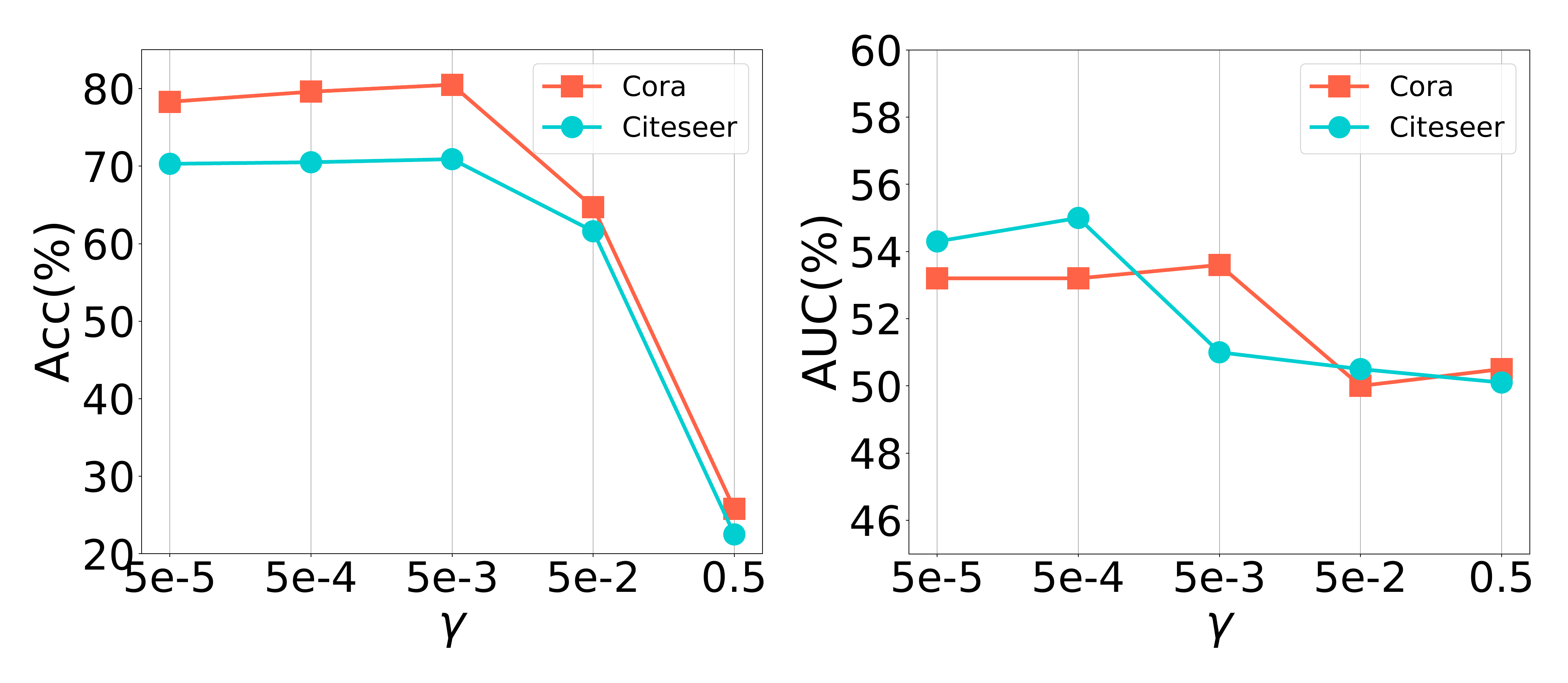}
    \vspace{-4mm}
    \caption{Performance of condensed graphs (left $\uparrow$) and MIA (right $\downarrow$) under different $\gamma$.}
    \label{fig:param_beta}
\vspace{-4mm}
\end{figure}

\subsection{Ablation Study}
\label{ab_study}
\noindent \textbf{Ablation study on condensed graph}.
We present the effects of different modules on condensed graphs, shown in Figure \ref{fig:ab_performance}. More results on other datasets are presented in the Appendix. $\textit{-com}$, $\textit{-ft}$,  $\textit{-ib}$, and $\textit{-st}$ means removing the communications, fine-tuning, local graph transformation, and self-training respectively. We can see that (1) One-step communications contribute more to relatively small graphs (Cora and Citeseer) than larger graphs (Ogbn-arxiv). The reason is that the smaller graphs suffer from more information losses (missing significant cross-client neighbors), leading to a cascading effect on the condensed graph \cite{DBLP:conf/icdm/Wang0PL022}. Similarly, self-training also plays a key role in smaller graphs since they need more labeled nodes to compensate for broken structures. These demonstrate the significance of integral local structures to the utility of FGC. (2) IB can improve the generalization ability of the condensed graph, which can be observed that removing IB will suffer from slight performance decay on almost all datasets. In fact, we instantiate IB by sampling $Z$ from the original feature distribution (Eq. (\ref{eq:pzxn})) and force it to be similar to the standard normal distribution, which avoids over-fitting with local data. (3) Fine-tuning with local data can achieve better personalization. Due to IB principles, the condensed graph captures the general knowledge of multiple subgraphs rather than specific local structure patterns. Thus, a few rounds of fine-tuning by local data can achieve better results.
%\end{enumerate}%[leftmargin=*]

\noindent \textbf{Ablation study on MIA}. The ablation study on MIA is depicted in Figure \ref{fig:ab_performance}. Results on other datasets are presented in the Appendix. We remove the $\textit{-ft}$ because it's conducted in the test phase and has no impact on MIA. We can see that the local graph transformation with IB has a significant defensive ability to MIA (up to 40\% MIA performance decay on Citeseer). As discussed before, the optimization of the IB objective in Eq. (\ref{eq:ib}) will minimize inference cost gain, making the MIA less effective. It can also be seen that one-step communication and self-training defend MIA to some degree. One-step communication can be seen as a feature augmentation and self-training aims to augment labels. These augmentations make the condensed graph fit to augmented data rather than original data thus alleviating the over-fitting, leading to a performance decay of MIA. 

\noindent \textbf{Performance during federated training}. We also present the training curve of the condensed graph and MIA w/ and w/o IB, shown in Figure \ref{fig:ab_ib_performance}. We can see that without IB ($\textit{FedGC-IB}$) the model converges slowly and also obtains lower performance than the model with IB ($\textit{FedGC}$). The reason for fast convergence lies in using self-training to label nodes before transforming graphs, which introduces more supervised signals to train the IB and consequently incorporates more accurate graph information into $Z$. The defense of MIA also benefits from the IB. FedGC consistently maintains a lower AUC score during the whole training process.

\subsection{Parameter Analysis}
\noindent \textbf{Analysis of different $\gamma$}. Figure \ref{fig:param_beta} presents the performance of condensed graph and MIA under different $\gamma$ of IB. A proper $\gamma$ can improve the generalization ability, but a large $\gamma$ will force $Z$ to follow specific distributions and perform poorly on downstream tasks. The MIA also has a similar trend. A smaller $\gamma$ causes the model to over-fit the training data, thereby enhancing MIA. On the contrary, a larger $\gamma$ makes $Z$ less similar to the original features, thus undermining the MIA. Overall, $\beta$ controls the trade-off between utility and privacy, which can be flexibly adjusted in reality. More parameter analysis can be found in the appendix.

\section{Conclusion}
In this paper, we investigate the novel problem of federated graph condensation. A general framework by matching weighted aggregated gradients is proposed to preserve the utility of condensed graphs. We also reveal the condensed graph will leak the membership privacy of local data and then propose a local graph transformation with information bottleneck principles to protect privacy. Empirical and theoretical analysis demonstrate the effectiveness of our method.

\section{Acknowledgments}
This work is supported in part by the National Key Research and Development Program of China (2023YFF0725103), the National Natural Science Foundation of China (U22B2038, 62192784), the JSPS KAKENHI JP23K24851, the JST PRESTO JPMJPR23P5, and the JST CREST JPMJCR21M2.

\bibliography{aaai25}

% \clearpage
% \input{checklist}

\clearpage
\appendix

\begin{algorithm}
    \renewcommand{\algorithmicrequire}{\textbf{Input:}}
    \renewcommand{\algorithmicensure}{\textbf{Output:}}
    \caption{FedGC} 
    \begin{algorithmic}[1]
        \REQUIRE  Subgraphs:$\mathcal{G}=\{G_i=(A_i,Z_i, Y_i)\}_{i=1}^M$; Client set: $\mathcal{U}=\{i\}_{i=1}^{M}$; Class set $\mathcal{C}$: $\mathcal{C}$
        \ENSURE Condensed graph $G^{\prime}=(A^{\prime}, X^{\prime}, Y^{\prime})$
        \STATE Clients perform one-time communications with the server to transmit neighbor aggregations
        \STATE Clients upload the number of nodes $n(c)$ for each $c \in \mathcal{C}$ 
        \STATE Server synthetizes $Y^{\prime}$ based on $\{n(c)\}_{c \in \mathcal{C}} $
        \STATE Server initializes parameter $W^{(1)}$, $W^{(2)}$, $\phi$ and distributes them to each client $i \in \mathcal{U}$
        \STATE $\phi_{st}$, $\{Y_i^{st}\}_{i \in \mathcal{U}}$ = \texttt{Self\_train}($\mathcal{G}$, $\mathcal{U}$)
        \STATE $\{Z_i\}_{i \in \mathcal{U}}$ = \texttt{Local\_graph\_transformation}($\mathcal{G}$, $\mathcal{U}$, $\phi_{st}$, $\{Y_i^{st}\}_{i \in \mathcal{U}}$)

        \WHILE{\textit{not converge}}
            \STATE Sever initializes $\theta$ and distributes it to each $i \in \mathcal{U}$
            \FOR{each client $i \in \mathcal{U}$}
                \STATE Sample subgraphs $G_{i,c}=(A_{i,c},Z_{i,c},Y_{i,c})$ for each class $c \in \mathcal{C}$
                \STATE Calculate the gradients $\{\nabla_{\theta} \mathcal{L}^{G_{i,c}}\}_{c \in C}$ based on $\{G_{i,c}\}_{c \in \mathcal{C}}$ and current $\theta$
                \STATE Upload $\{\nabla_{\theta} \mathcal{L}^{G_{i,c}}\}_{c \in C}$ to the server
            \ENDFOR
            \STATE Server aggregates $\{\nabla_{\theta} \mathcal{L}^{G_{i,c}}\}_{c \in C}$ and calculates the matching loss based on Eq.(\ref{eq:fedgc1})
            \STATE Server updates parameters $X^{\prime}$ and $\Phi$ of $G^{\prime}$
        \ENDWHILE 
        \STATE Server synthesizes $G^{\prime}$ using $X^{\prime}$ and $\Phi$
        \STATE \textbf{return}  $G^{\prime}$
        
        ~\\
        \STATE $\textbf{Function}$ 
        $\texttt{Local\_graph\_transformation}$ ($\mathcal{G}$, $\mathcal{U}$, $\phi_{st}$, $\{Y_i^{st}\}_{i \in \mathcal{U}}$):
        \STATE Initialize $\phi$ with $\phi_{st}$ and update $Y_i$ with $Y_i^{st}$ for each $i \in \mathcal{U}$
        \WHILE{\textit{not converge}} 
            \FOR{each client $i \in \mathcal{U}$}
                \STATE Calculate $Z_i$ based on Eq.(\ref{eq:pzxn})
                \STATE Update best $Z^{*}_i$ based on valid data set
                \STATE Update $W^{(1)}$, $W^{(2)}$ by minimizing $\mathcal{L}_{IB}$ in Eq.(\ref{eq:gt}) 
            \ENDFOR
        \ENDWHILE
        \STATE \textbf{return}  $\{Z^{*}_i\}_{i \in \mathcal{U}}$
    
        ~\\
        %HGNN_for_rec
        \STATE $\textbf{Function}$ $\texttt{Self\_train}$ ($\mathcal{G}$, $\mathcal{U}$):
        \WHILE{\textit{not converge}} 
            \FOR{each client $i \in \mathcal{U}$}
                \STATE Update best model $\phi_{st}$ based on valid data set 
                \STATE Calculate gradients $\nabla_{\phi} \mathcal{L}^{G_{i}}$ base on $G_i$ 
                \STATE Upload $\nabla_{\phi} \mathcal{L}^{G_{i}}$ to the server
            \ENDFOR
            \STATE Server aggregates $\nabla_{\phi} \mathcal{L}^{G_{i}}$ and updates $\phi$
            \STATE Server distributes $\phi$ to each $i \in \mathcal{U}$
        \ENDWHILE
        \FOR{each client $i \in \mathcal{U}$}
                \STATE Label unlabeled node set $\{v|v \in V_i/V_i^{train}\}$ based on model with $\phi_{st}$ and obtain new labeled set $Y_i^{st}$
        \ENDFOR
        \STATE \textbf{return} $\phi_{st}$, $\{Y_i^{st}\}_{i \in \mathcal{U}}$
    \end{algorithmic}
\label{alg1}
\end{algorithm}
%$\hspace{0.4cm}

\section{Proofs}
Let $\mathcal{G}^{\prime}=(A^{\prime},X^{\prime})$, $\mathcal{G}_i=(A_i,X_i)$, $\mathcal{G}_i^t=(A_i,Z_i)$ denote condensed graph, client $i$'s original graph and transformed graph respectively. Assuming $Y_i \rightarrow   \mathcal{G}_i \rightarrow \mathcal{G}_i^t \rightarrow \mathcal{G}^{\prime}$ form a Markov chain, the attack aims to find a graph $g$ to match $\mathcal{G}_i$ with an inference cost $C(\mathcal{G}_i, g)$. Part of the proofs is motivated by \cite{DBLP:conf/itw/MakhdoumiSFM14}.  
 
\subsection{Proof of Lemma 1}

\noindent \textbf{Lemma 1.} If $L = \sup|C(\mathcal{G}_i, g)| < \infty$ is satisfied, then $\Delta C = c_{\mathcal{G}_i}^* - \mathbb{E}_{P_{\mathcal{G}^{\prime}}} \left[ c_{\mathcal{G}^{\prime}}^* \right] \leq 2\sqrt{2}L \sqrt{\text{I}(\mathcal{G}_i; \mathcal{G}^{\prime} )}$.

%= \min_{g \in P_{\mathcal{G}_i}} \mathbb{E}_{P_{\mathcal{G}_i}} [C({\mathcal{G}_i}, g)] 

% = \mathbb{E}_{P_{\mathcal{G}^{\prime}}} [\min_{g \in P_{\mathcal{G}_i}} \mathbb{E}_{P_{\mathcal{G}_i|\mathcal{G}^{\prime}}} \left[ C(\mathcal{G}_i, g) \mid \mathcal{G}^{\prime} = g^{\prime} \right]] 

\begin{proof} Let 
$g^{*}_{\mathcal{G}_i} 
= \arg \min_{g \in P_{\mathcal{G}_i}} \mathbb{E}_{P_{\mathcal{G}_i}} [C({\mathcal{G}_i}, g)]$ 
and 
$g^{*}_{g^{\prime}} 
= \arg \min_{g \in P_{\mathcal{G}_i}} \mathbb{E}_{P_{\mathcal{G}_i|\mathcal{G}^{\prime}}} \left[ C(\mathcal{G}_i, g) \mid \mathcal{G}^{\prime} = g^{\prime} \right]] $ denote the inferred graph w/o and w/ observing $\mathcal{G}^{\prime}$ respectively, we have 
$c_{\mathcal{G}_i}^* 
= \mathbb{E}_{P_{\mathcal{G}_i}} [C({\mathcal{G}_i}, g^{*}_{\mathcal{G}_i})]$ 
and 
$\mathbb{E}_{P_{\mathcal{G}^{\prime}}}\left[c^*_{\mathcal{G}^{\prime}}\right] 
= \mathbb{E}_{P_{\mathcal{G}^{\prime}}} [\mathbb{E}_{P_{\mathcal{G}_i|\mathcal{G}^{\prime}}} \left[ C(\mathcal{G}_i, g^{*}_{g^{\prime}}) \mid \mathcal{G}^{\prime} = g^{\prime} \right]]$. Thus,

\begin{equation}
    \begin{aligned}
    &\Delta C \\
    &= \mathbb{E}_{P_{\mathcal{G}_i}} [C({\mathcal{G}_i}, g^{*}_{\mathcal{G}_i})] - \mathbb{E}_{P_{\mathcal{G}^{\prime}}} [\mathbb{E}_{P_{\mathcal{G}_i|\mathcal{G}^{\prime}}} \left[ C(\mathcal{G}_i, g^{*}_{g^{\prime}}) \mid \mathcal{G}^{\prime} = g^{\prime} \right]] \\
    &=\mathbb{E}_{P_{\mathcal{G}^{\prime}}} [\mathbb{E}_{P_{\mathcal{G}_i|\mathcal{G}^{\prime}}}[C(\mathcal{G}_i, g^{*}_{\mathcal{G}_i}) - C(\mathcal{G}_i, g^{*}_{g^{\prime}}) | \mathcal{G}^{\prime} = g^{\prime}]].
    \end{aligned}
\label{eq:1}
\end{equation}
We can also derive that 
\begin{equation}
    \begin{aligned}
        &\mathbb{E}_{P_{\mathcal{G}_i|\mathcal{G}^{\prime}}}[C(\mathcal{G}_i, g^{*}_{\mathcal{G}_i}) - C(\mathcal{G}_i, g^{*}_{g^{\prime}}) | \mathcal{G}^{\prime} = g^{\prime}]\\
        &= \sum_{g_{i}} p(g_{i}|g^{\prime})[C(g_{i}, g^{*}_{\mathcal{G}_i}) - C(g_{i}, g_{g^{\prime}}^*)] \\
        &= \sum_{g_{i}} (p(g_{i}|g^{\prime}) - p(g_{i}) + p(g_{i}))[C(g_{i}, g^{*}_{\mathcal{G}_i}) - C(g_{i}, g_{g^{\prime}}^*)] \\
        &= \sum_{g_{i}} (p(g_{i}|g^{\prime}) - p(g_{i}))[C(g_{i}, g^{*}_{\mathcal{G}_i}) - C(g_{i}, g_{g^{\prime}}^*)] \\
        &+ \sum_{g_{i}} p(g_{i})[C(g_{i}, g^{*}_{\mathcal{G}_i}) - C(g_{i}, g_{g^{\prime}}^*)] \\
        &\leq 2L \sum_{g_{i}} |p(g_{i}|g^{\prime}) - p(g_{i})| \\
        &+ (\mathbb{E}_{P_{\mathcal{G}_i}}[C(\mathcal{G}_i, g^{*}_{\mathcal{G}_i})] -\mathbb{E}_{P_{\mathcal{G}_i}}[C(\mathcal{G}_i, g_{g^{\prime}}^*)]), \\
        &\leq 2L \sum_{g_i} |p(g_{i}|g^{\prime}) - p(g_{i})|, \\
        &\leq 4L \|P_{\mathcal{G}_i|\mathcal{G}^{\prime}=g^{\prime}} - P_{\mathcal{G}_i}\|_{TV} \\
        &\leq 4L \sqrt{\frac{1}{2}D(P_{\mathcal{G}_i|\mathcal{G}^{\prime}=g^{\prime}} \| P_{\mathcal{G}_i})},
    \end{aligned}
\label{eq:2}
\end{equation}
where $\|\cdot\|_{TV}$ denotes the total variation (TV) distance and $D(\cdot \| \cdot)$ is the KL divergence. The last inequality is derived from the Pinsker's inequality. Combining Eq. (\ref{eq:1}) and Eq. (\ref{eq:2}), we have 
\begin{equation}
    \begin{aligned}
    \Delta C 
    &\leq 2\sqrt{2}L \mathbb{E}_{P_{\mathcal{G}^{\prime}}} \left[ \sqrt{D(P_{\mathcal{G}_i|\mathcal{G}^{\prime}=g^{\prime}} \| P_{\mathcal{G}_i})} \right]\\
    &\leq 2\sqrt{2}L \sqrt{\text{I}(\mathcal{G}_i; \mathcal{G}^{\prime})}.
    \end{aligned}
\label{eq:3}
\end{equation}
The last inequality establishes because the square root function is concave and thus Jensen's inequality can be applied. 

\end{proof}

\subsection{Proof of Lemma 2}

\noindent \textbf{Lemma 2.} Minimizing $\text{I}(\mathcal{G}_i; \mathcal{G}_i^t)$ is equivalent to minimizing $\text{I}(A_i, X_i; Z_i)$.

\begin{proof} 
We assume the transformed feature $Z_i$ is independent of the adjacency matrix $A_i$, then we have 
    \begin{equation}
    \begin{aligned}
    &\text{I}(\mathcal{G}_i; \mathcal{G}_i^t) =\text{I}(A_i,X_i; A_i,Z_i) \\
    &=\text{I}(X_i,A_i;Z_i) + \text{I}(X_i,A_i;A_i|Z_i)\\
    &=\text{I}(X_i,A_i;Z_i) + H(A_i|Z_i)-H(A_i|X_i,A_i,Z_i)\\
    &=\text{I}(X_i,A_i;Z_i) + H(A_i|Z_i)\\
    &=\text{I}(X_i,A_i;Z_i) + H(A_i).
    \end{aligned}
    \end{equation}
$H(A_i)$ can be seen as a constant since the adjacency matrix $A_i$ is unchanged during the whole local graph transformation process. Consequently, minimizing $\text{I}(\mathcal{G}_i; \mathcal{G}_i^t)$ is equivalent to minimizing $\text{I}(A_i, X_i; Z_i)$.
\end{proof}

\section{Algorithms}
\label{alg}
 The algorithm of FedGC is shown in Algorithm \ref{alg1}. Given multiple subgraphs $\{G_i=(X_i,A_i,Y_i)\}_{i=1}^M$ stored in the corresponding client, before federated graph condensation, each client needs to perform (1) one-time communications with server to obtain aggregated neighbor representations (line 2). (2) collaboratively training a global GNN model $\phi_{st}$ to labeled local nodes (lines 27-43) (3) locally training a graph transformation module to obtain a new subgraph $G_i^t=(A_i,Z_i,Y_i)$ (lines 17-26). After that, federated graph condensation is conducted based on multiple transformed subgraphs until model convergence (lines 5-16). During the test phase, the condensed graph $G^{\prime}$ is first utilized to train a GNN model, and then the GNN model will be fine-tuned by locally transformed graphs. Finally, the model will be used to test local data. Since we focus on obtaining a condensed graph, the test phase is omitted in Algorithm \ref{alg1} for clarity.

\begin{table*}[t]
\renewcommand{\arraystretch}{1.4} %控制行高
  \centering
  \small
  % \resizebox{\columnwidth}{!}{
    \begin{tabular}{lccccc}
\toprule
Dataset & \#Nodes & \#Edges & \#Classes & \#Features & Training/Validation/Test \\
\midrule
Cora & 2,708 & 5,429 & 7 & 1,433 & 140/500/1000 \\
Citeeseer & 3,327 & 4,732 & 6 & 3,703 & 120/500/1000 \\
Flickr & 89,250 & 899,756 & 7 & 500 & 44,625/22,312/22,313 \\
Ogbn-arxiv & 169,343 & 1,166,243 & 40 & 128 & 90,941/29,799/48,603 \\
Reddit & 232,965 & 57,307,946 & 210 & 602 & 153,932/23,699/55,334 \\
\bottomrule
    \end{tabular}%
\caption{Dataset statistics.}
\label{tab:data}%

\end{table*}%

% \mathcal{G}_i
% \mathcal{G}^{\prime}
% g^{\prime}
% g^{*}_{\mathcal{G}_i}
% g^{*}_{g^{\prime}}
% g_{i}

\section{Datasets}
\label{data}
Following \cite{DBLP:conf/iclr/JinZZLTS22, DBLP:conf/kdd/JinTJLZTY22, DBLP:conf/nips/ZhengZCN0P23}, we use five datasets to evaluate our model. \textbf{Cora}, \textbf{Citeseer}, and \textbf{Ogbn-arxiv} are citation networks, where nodes represent the papers and edges denote citation relations. \textbf{Flickr} is a graph used for image classification, where the node represents the image with descriptions as node attributes and the edges connect nodes that share common properties. \textbf{Reddit} is a large-scale network of online discussion forums where the node is the post and connecting posts if the same user comments on both. Table \ref{tab:data} presents the detailed statistics of the dataset. We use the public splits for all the datasets.

\section{Baselines}
\label{baseline}
The details of baselines are summarized as follows:

\begin{itemize}%[leftmargin=*]

\item \textbf{GCN} \cite{DBLP:conf/iclr/KipfW17} is a typical semi-supervised graph convolutional network that defines the graph convolution in light of spectral analysis.

\item \textbf{GCOND} \cite{DBLP:conf/iclr/JinZZLTS22} is the first work adapting gradient matching for graph condensation.

\item \textbf{DosCond} \cite{DBLP:conf/kdd/JinTJLZTY22} develops an efficient one-step gradient matching strategy for graph condensation.

\item \textbf{SFGC} \cite{DBLP:conf/nips/ZhengZCN0P23} is based on training trajectory matching and is the current state-of-the-art method.

\item \textbf{BDS-GCN} \cite{DBLP:conf/mlsys/WanLLKL22} uses a boundary sampling strategy to sample cross-client edges in each training round and thus may leak data privacy.

\item \textbf{FedSage+} \cite{DBLP:conf/icml/BaekJJYH23} trains missing neighborhood and feature generators to predict the graph structure and associated embeddings but sacrificing computational efficiency.

\item \textbf{FedGCN} \cite{DBLP:conf/nips/YaoJRJ23} is the state-of-the-art federated graph learning method based on one-step transmitting neighbor aggregations among clients.

\item \textbf{PL} \cite{lee2013pseudo} uses self-training to obtain pseudo labels and then perform GCN based on all labeled nodes. We first collaboratively train a GCN model to label local nodes and then perform FGC on all labeled nodes.

\item \textbf{Reg} \cite{DBLP:conf/ccs/NasrSH18} introduces an adversarial regularizer to train the model for maximum privacy-preserving against the strongest MIA. Considering this is a model-based defense method, we add the regularizer to the local model for adjusting the uploaded gradients. In this way, by matching uploaded gradients the condensed graph is optimized to be robust to MIA.

\item \textbf{LDP} \cite{DBLP:conf/isca/ChoiTVHK18} adds noises to the data to guarantee the Local Differential Privacy (LDP). Following \cite{wu2022federated}, we add Laplace noises to the uploaded gradients with a privacy budget $\epsilon=1$ to satisfy LDP.

\end{itemize}

\section{Implementation Details}
The condensation ratio $r$ is defined as the proportion of condensed nodes to original nodes \cite{DBLP:conf/iclr/JinZZLTS22}. Following \cite{DBLP:conf/iclr/JinZZLTS22, DBLP:conf/nips/ZhengZCN0P23} ,for small dataset Cora and Citeseer, we choose $r$ as \{1.3\%, 2.6\%, 5.2\%\} and \{0.9\%, 1.8\%, 3.6\%\} respectively. For large-scale datasets Ogbn-arxiv, Flickr, and Reddit, we choose $r$ as \{0.05\%, 0.25\%, 0.5\%\}, \{0.1\%, 0.5\%, 1\%\} and \{0.05\%, 0.1\%, 0.2\%\} respectively. For all the baselines, the node features are randomly initialized and the hidden dimension is set to 256. For FedGC, we utilize a 2-layer GCN \cite{DBLP:conf/iclr/KipfW17} in the self-training phrase and test phrase. For the condensation methods, we use a 2-layer GCN  as the backbone (3-layer SGC \cite{DBLP:conf/icml/WuSZFYW19} for Ogbn-arxiv). We use Mean Square Error (MSE) as the distance function of gradient matching for all datasets, except for using the distance from \cite{DBLP:conf/iclr/JinZZLTS22} in Ogbn-arxiv:

\begin{equation} 
    D(g^1, g^2) = \sum_{i=1}^{d}(1-\frac{g^1_i\cdot g^2_i}{\left \|g^1_i\right \| \left \|g^2_i\right \|}).
    \label{eq:distance}
\end{equation}
where $g^1$ and $g^2$ are two $d$-dimensional gradients and $g_i$ is the $i$-th column vector of the gradient. The detailed calculation of condensed graph structure $A^{\prime}$ is:

\begin{equation} 
    A_{i,j}^{\prime} = \sigma (\frac{\text{MLP}_\Phi ([x_i^{\prime} \left\vert \right\vert x_j^{\prime}])+\text{MLP}_\Phi([x_j^{\prime} \left\vert \right\vert x_i^{\prime}])}{2})
    \label{eq:model_ax}
\end{equation}
where $[\cdot \left\vert \right\vert \cdot]$ denotes concatenation operation and we force the condensed graph to be symmetric by making $A_{i,j}^{\prime} = A_{j,i}^{\prime}$. Also, we can change it to an asymmetric graph by using $A_{i,j}^{\prime}=\sigma(\text{MLP}_{\Phi}(x_i^{\prime} \left\vert \right\vert x_j^{\prime}))$. 

For MIA, following \cite{DBLP:journals/corr/abs-2102-05429}, we assume that the attacker possesses a subgraph of the dataset with a small set of labels to form the shadow dataset. The nodes used for training and testing shadow models have no overlaps with the original train and test nodes. They are all randomly sampled with the same size. The attack model is a 3-layer MLP and the shadow and target model are 2-layer GCN. For all the baselines, we tune other hyper-parameters based on valid datasets to report the best performance.

\section{More Experimental Results}

\subsection{Generalizability of condensed graphs on more datasets}
Table \ref{app:generalization} presents the generalizability of condensed graphs in Flickr and Reddit datasets. We can see that FedGC outperforms GCOND and SFGC w.r.t all the GNN architectures and even the simple MLP. The improvements are significant, demonstrating the good generalizability of the condensed graph by FedGC, especially in large-scale graph datasets.

\subsection{Ablation study on more datasets}
\begin{figure}[t]
    \centering
    \includegraphics[scale=0.23]{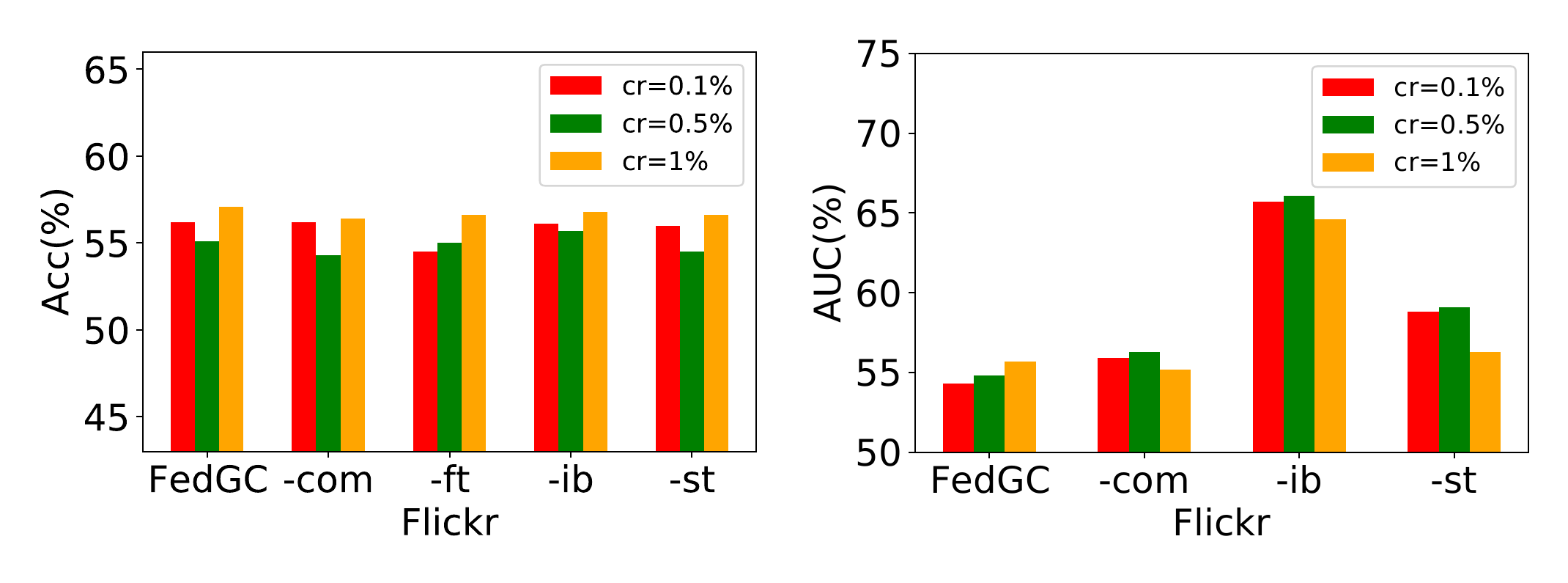}
    \vspace{-4mm}
    \caption{Ablation study for the performance of condensed graph (left $\uparrow$) and MIA (right $\downarrow$).}
    \label{fig:ab_performance_appendix}
\end{figure}

The ablation study for the performance on Flickr dataset is depicted in Figure \ref{fig:ab_performance_appendix}. It can be observed that all the components in FedGC contribute to the utility of the condensed graph and the proposed local graph transformation with IB principles can largely reduce the MIA risks.

\begin{table}[t]
\centering
\setlength\tabcolsep{1mm}
\footnotesize
\begin{tabular}
{@{}cccccccccc@{}}
\toprule
\multicolumn{1}{l}{}                                                                 & Methods & MLP     & \multicolumn{1}{m{0.57cm}<{\centering}}{AP.}  & Cheby  & GCN    & SAGE   & SGC    & Avg.   \\ \midrule

\multirow{3}{*}{\begin{tabular}[c]{@{}c@{}}Flickr\\      $0.5\%$\end{tabular}}       & GCOND    & $43.3$      & $49.7$ & $46.2$ & $49.7$ & $49.4$ & $46.0$ & $48.2$ \\
& SFGC    & $49.4$      & $46.7$ & $48.0$ & $49.5$ & $49.3$ & $47.2$ & $48.2$ \\
& \text{FedGC} & $47.2$  & $55.1$ & $54.3$ & $55.1$ & $55.7$ & $53.5$ & $54.7$ \\\midrule
\multirow{3}{*}{\begin{tabular}[c]{@{}c@{}}Reddit\\       $0.1\%$\end{tabular}}      & GCOND     & $44.2$       & $87.7$ & $76.1$ & $89.8$ & $89.0$ & $91.3$ & $86.8$ \\
& SFGC     & $90.7$       & $89.7$ & $91.2$ & $90.1$ & $91.2$ & $90.6$ & $90.6$ \\
& \text{FedGC}  & $91.9$ & $94.5$ & $94.5$ & $94.3$ & $94.3$   & $94.5$ & $94.4$ \\
\bottomrule

\end{tabular}
\caption{Generalizability (\%) of condensed graphs by FedGC. \textbf{AP.}:APPNP. \textbf{Avg.}: the average test accuracy across different architectures (excluding MLP).}
\label{app:generalization}
\end{table}

\subsection{Parameter analysis on different number of clients}
\begin{figure}[t]
    \centering
    \includegraphics[scale=0.12]{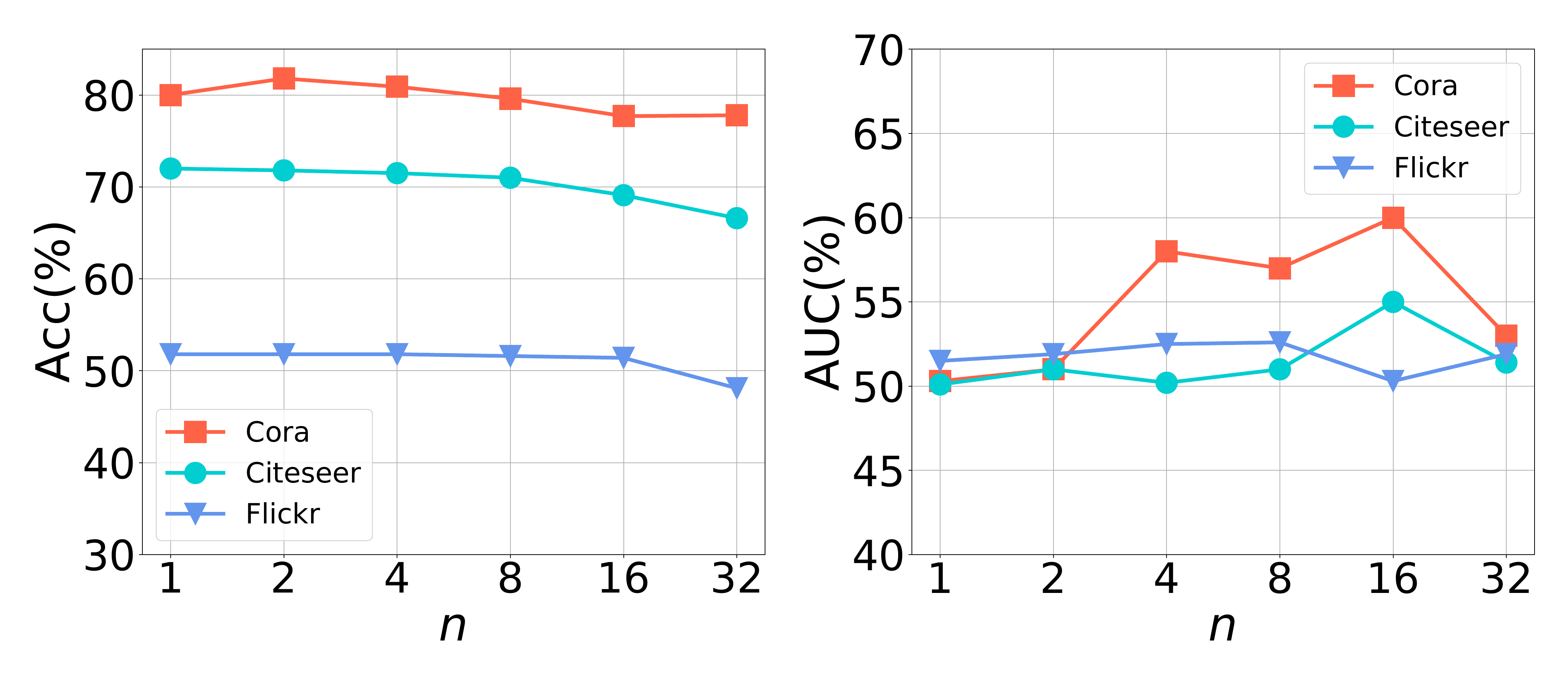}
    \caption{Performance of FedGC on different client number $n$ (left: performance of condensed graph ($\uparrow$), right: performance of MIA ($\downarrow$)).}
    \label{fig:param_ntrainer}
\end{figure}

The performance of FedGC under different numbers of clients is shown in Figure \ref{fig:param_ntrainer}. The stable performance of the condensed graph demonstrates that FedGC presents good robustness for various numbers of clients. The performance of MIA is also stable on Citeseer and Flickr but has an increased trend on Cora. Considering the slight decrease in performance of the condensed graph when $n$ increases, we assume the reason is that the larger $n$ makes fewer train data in each client, thus the condensed graph may easily over-fit the local train data, leading to a better MIA. Self-training in the condensed phrase is a promising solution. However, it will introduce more communication and computation costs. We leave this as our future work.

\subsection{Converges analysis}
\begin{figure}[t]
    \centering
    \includegraphics[scale=0.115]{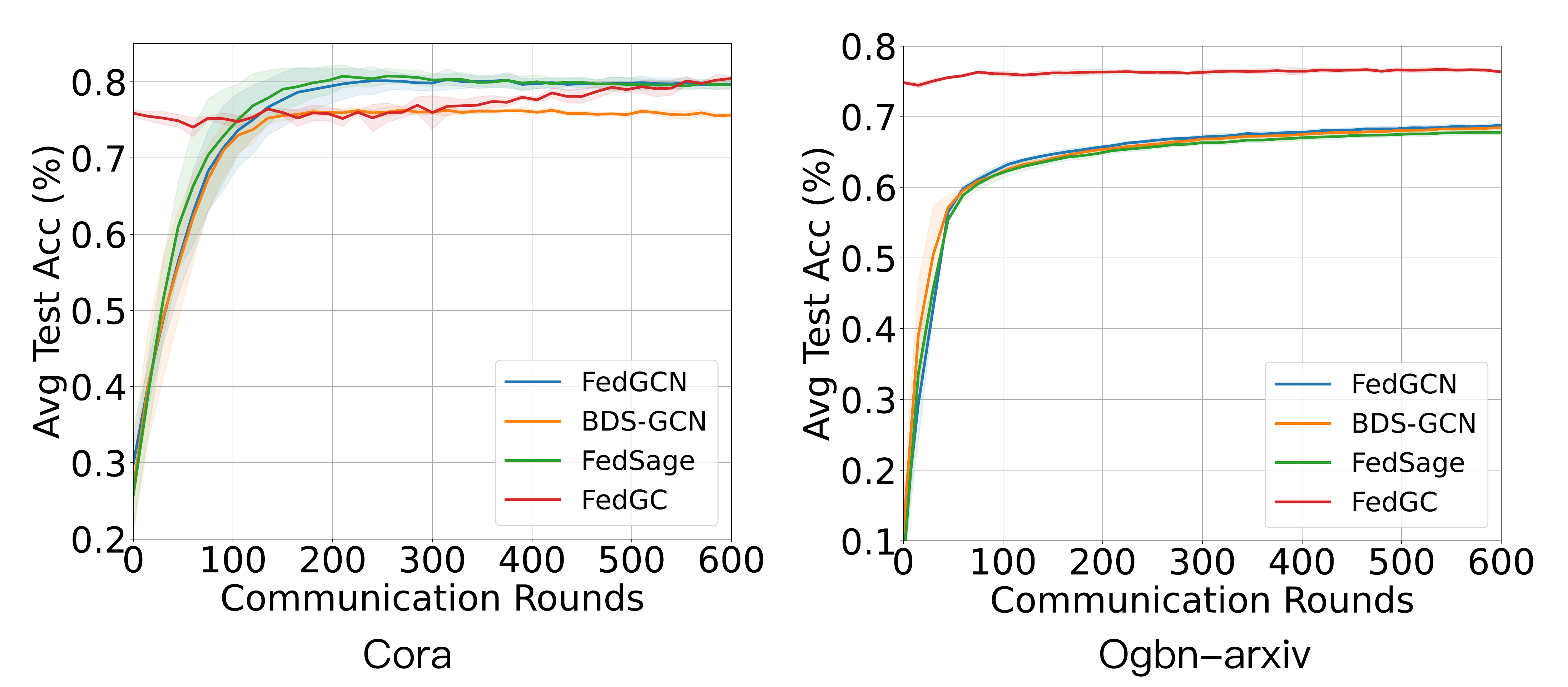}
    \caption{Convergence curve of FedGC against existing FGL baselines.}
    \label{fig:convergence}
\end{figure}

As can be seen in Figure \ref{fig:convergence}, compared to Federated Graph Learning (FGL) baselines, our framework converges much faster with higher accuracy on Ogbn-arxiv, demonstrating its superiority in handling real-world large-scale graphs. 

\subsection{Running time}

Following \cite{DBLP:conf/iclr/JinZZLTS22}, we report the running time of FedGC under different condensation rates. Concretely, we vary the rates in the range of \{1.3\%, 2.6\%, 5.2\%\} on Cora and \{0.05\%, 0.25\%, 0.5\%\} on Ogbn-arxiv. For Ogbn-arxiv, we run 10 epochs (communication rounds) on two NVIDIA GeForce RTX 2080 Ti GPU with ray \cite{DBLP:conf/osdi/MoritzNWTLLEYPJ18} for parallelization. We use one NVIDIA GeForce RTX 2080 Ti GPU with no parallelization for Cora. The results are reported in Table 
\ref{tab:runtime}. The entire condensation process (1000 epochs) for generating a condensed graph with $r=0.25\%$ of the Ogbn-arxiv dataset takes approximately 3.6 hours. In addition, we record the running time in the self-train stage and local graph transformation stage. It takes about 1.8 hours and 17 minutes respectively on Ogbn-arxiv dataset. The running time is acceptable considering the huge benefits of condensed graphs. For example, once we obtain the condensed graph, it only takes about 1 minute to fine-tune with local data to achieve high performance on Ogbn-arxiv.

\begin{table}[h]
\scriptsize
\setlength\tabcolsep{1.5mm}
\centering
\begin{tabular}{ccccccccc}
\toprule
r & 0.05\%& 0.25\% &0.5\% &\textbar &r&1.3\%&2.6\%&5.2\% \\
\midrule

Ogbn-arxiv & 2.3min & 2.2min & 2.2min & \textbar &  
Cora & 3.2s & 3.4s & 3.6s  \\
\bottomrule
\end{tabular}
\caption{Running time of FedGC for 10 epochs in the condensation stage.}
\label{tab:runtime}
\end{table}

\section{Related work}
\textbf{Dataset condensation}. Due to the prohibitive costs of training models with large-scale datasets, dataset condensation (DC) has emerged to synthesize a smaller dataset by distilling the knowledge of a given large dataset. Subsequently, the condensed dataset can serve as a substitute for the large dataset to conduct downstream tasks, such as hyper-parameter tuning and neural architecture search (NAS). Typically, dataset distillation (DD) \cite{DBLP:journals/corr/abs-1811-10959} firstly achieves this by a learning-to-learn framework with bi-level optimization. To alleviate the high computational overhead brought by the unrolled computation graph, DC \cite{DBLP:conf/iclr/ZhaoMB21, DBLP:conf/icml/ZhaoB21} is proposed to match the gradients between the original graph and the condensed graph. Similar idea is also adopted to match the distributions \cite{DBLP:conf/wacv/ZhaoB23} and training trajectories \cite{DBLP:conf/cvpr/Cazenavette00EZ22b}. Extending to the graph data, GCOND \cite{DBLP:conf/iclr/JinZZLTS22} condenses a large graph into a smaller one for node classifications based on gradient matching scheme. DosCond \cite{DBLP:conf/kdd/JinTJLZTY22} further proposes a one-step gradient matching algorithm to avoid nested optimization of GCOND. It is also compatible with graph classification tasks. Other works attempt to match the distributions \cite{DBLP:journals/corr/abs-2206-13697} and training trajectories \cite{DBLP:conf/nips/ZhengZCN0P23} of graph data and achieve remarkable results. Despite their success, they focus on the graph condensation under centralized data storage, regardless of restrained conditions with distributed data or local privacy-preserving requirements.

\noindent \textbf{Federated graph learning}. 
Federated learning (FL) \cite{DBLP:conf/aistats/McMahanMRHA17, DBLP:series/lncs/12500} has emerged to endow the ability of collaboratively learning a global model with distributed data while protecting data privacy. Federated graph learning (FGL) extends the FL to the graph domain and can be divided into inter-graph and intra-graph learning based on the data distributed type \cite{DBLP:journals/corr/abs-2105-11099}. In inter-graph learning, each client owns multiple graphs and aims to train a global model with other clients' cooperation for graph-level tasks \cite{DBLP:conf/nips/XieMXY21}, such as molecular property prediction \cite{DBLP:journals/patterns/ZhuLW22}. In contrast, intra-graph learning assumes each client only owns a subgraph of the whole graph, which therefore mainly focuses on recovering missing cross-client edges and neighbors. BDS-GCN \cite{DBLP:conf/mlsys/WanLLKL22} proposes sampling cross-client neighbors but violates other clients' privacy. Instead, FedSage+ \cite{DBLP:conf/nips/ZhangYLSY21} generates missing neighbors locally by training linear generators. However, it relies on communications between clients in each federated training round. To reduce communication costs, FedGCN \cite{DBLP:conf/nips/YaoJRJ23} exchanges neighbor aggregations through homomorphic encryption in one pre-training step, largely speeding up the federated training process and also protecting local privacy. Some other intra-graph learning works are dedicated to tackling the heterogeneity issues among clients \cite{DBLP:conf/icml/BaekJJYH23, DBLP:conf/aaai/TanLL00Z23}. They model the similarities or share common knowledge between clients to alleviate the performance degradation caused by data heterogeneity. All these existing federated graph learning methods require burdensome federated training processes, which even need to be repeated multiple times in some downstream tasks, such as hyper-parameter tuning and neural architecture search. In light of this limitation, we focus on intra-graph learning and study federated graph condensation to obtain a global small graph for facilitating downstream tasks.

\noindent \textbf{Membership inference attack}. 
Membership inference attack (MIA)\cite{DBLP:conf/ndss/Salem0HBF019, DBLP:conf/ccs/NasrSH18, DBLP:conf/uss/Carlini0EKS19, DBLP:conf/ccs/ChenL0023, DBLP:conf/ccs/ZhangRWRCHZ21} is a privacy attack enabling an attacker to illicitly obtain private or sensitive data from a learning model, by discerning the use of a specific sample as training data. Shadow training is generally used to train the attack model for MIA \cite{DBLP:conf/sp/ShokriSSS17, DBLP:conf/ndss/Salem0HBF019}, in the case that the training data with some background knowledge, such as full output knowledge, partial output knowledge, and label-only knowledge, is queried by the attacker to be the target posterior, the attacker trains multiple shadow models to simulate the target model and uses multiple attack models to attack. There may be differences in the setting of adversarial shadowing model learning attacks, the target victim model access may be black-box or white-box for the attacker\cite{DBLP:conf/icdm/LiYZ18}. In a white-box setting, the adversary has complete access to the target victim model, leveraging its entire structure to create adversarial examples and attack models, typically resulting in a near 100\% attack success rate. When the adversary has black-box access, which is much more practical, it can only query the target model without knowledge of the model’s structure or parameters (e.g., weights, gradients) or underlying assumptions regarding the model architecture. Several works leverage data augmentations \cite{DBLP:conf/cvpr/KahlaCJJ22} and confident scores \cite{DBLP:conf/ndss/Salem0HBF019} from the victim model’s output to readily reconstruct local sensitive data, posing a threat to vulnerable learning models on the server. Recently, researches \cite{DBLP:conf/tpsisa/OlatunjiNK21, DBLP:journals/corr/abs-2102-05429} have indicated that GNN is vulnerable to MIA, and the graph properties such as connectivity information encoded by the GNN inside the model structure makes it more vulnerable to such inference attacks. Their attacks are conducted in a restricted scenario, using a target node’s subgraph to query the target model to obtain the input to their attack model.

\noindent \textbf{Graph Neural Network}. 
Graph Neural Networks (GNNs) \cite{DBLP:journals/tnn/WuPCLZY21, DBLP:journals/corr/abs-2408-08685, DBLP:conf/iclr/XuHLJ19} have emerged as a powerful tool for modeling structure data and have been applied in various fields such as recommendation systems \cite{DBLP:conf/kdd/FanZHSHML19}, bioinformatics \cite{DBLP:conf/nips/FoutBSB17}, and cybersecurity \cite{DBLP:conf/ccs/QinXL23}. As a most popular GNN, GCN \cite{DBLP:conf/iclr/KipfW17} simplifies the convolution process on graphs by using an efficient layer-wise propagation rule. SGC \cite{DBLP:conf/icml/WuSZFYW19} removes the nonlinear activation function in GCN to speed up the training process and also keep the competitive performance. GAT \cite{DBLP:conf/iclr/VelickovicCCRLB18} further incorporates attention mechanisms to weigh the influence of nodes during the feature aggregation phase. GraphSage \cite{DBLP:conf/nips/HamiltonYL17} enables inductive learning on graphs by sampling and aggregating features from local neighborhoods. To solve the over-smoothing problem of GNNs, APPNP \cite{DBLP:conf/iclr/KlicperaBG19} is proposed to decouple the feature transformation from the propagation with an adaptation of personalized PageRank. Besides above mentioned spatial GNNs, there are some GNNs that leverage the spectrum of graph Laplacian to perform convolutions in the spectral domain \cite{DBLP:conf/iclr/BoSWL23, DBLP:conf/aaai/BoWSS21}. For example, ChebyNet \cite{DBLP:conf/nips/DefferrardBV16} is an implementation method of CNNs on graphs with fast localized spectral filtering. In this paper, we use GNNs as a backbone model for graph condensation and test the performance of the condensed graph under different GNN architectures.

\end{document}